\documentclass{article}
\usepackage{amssymb}

\usepackage{amsmath}
\usepackage{graphicx}
\usepackage{bm}
\usepackage{enumitem}
\usepackage{caption}
\usepackage{subcaption}
\usepackage{wrapfig}
\usepackage[font=footnotesize]{caption}  

\usepackage[preprint]{corl_2025} 

\title{Crossing the Human-Robot Embodiment Gap with Sim-to-Real RL using One Human Demonstration}

\author{
  Tyler Ga Wei Lum$^{*}$, Olivia Y. Lee$^{*}$, C. Karen Liu, Jeannette Bohg \\
  Stanford University \\
  \texttt{\{tylerlum, oliviayl, ckliu38, bohg\}@stanford.edu} \\
  \small $^{*}$ Equal Contribution\\
}

\begin{document}
\maketitle

\vspace{-0.5cm}
\begin{abstract}
    Teaching robots dexterous manipulation skills often requires collecting hundreds of demonstrations using wearables or teleoperation, a process that is challenging to scale. Videos of human-object interactions are easier to collect and scale, but leveraging them directly for robot learning is difficult due to the lack of explicit action labels and human-robot embodiment differences. We propose \textsc{Human2Sim2Robot}, a novel real-to-sim-to-real framework for training dexterous manipulation policies using only one RGB-D video of a human demonstrating a task. Our method utilizes reinforcement learning (RL) in simulation to cross the embodiment gap without relying on wearables, teleoperation, or large-scale data collection. From the video, we extract: (1) the object pose trajectory to define an object-centric, embodiment-agnostic reward, and (2) the pre-manipulation hand pose to initialize and guide exploration during RL training. These components enable effective policy learning without any task-specific reward tuning. In the single human demo regime, \textsc{Human2Sim2Robot} outperforms object-aware replay by over 55\% and imitation learning by over 68\% on grasping, non-prehensile manipulation, and multi-step tasks. Website: \href{human2sim2robot.github.io}{human2sim2robot.github.io}
\end{abstract}


\section{Introduction}

\begin{wrapfigure}{r}{0.55\textwidth}
\vspace{-0.4cm}
\centering
\includegraphics[width=\linewidth]{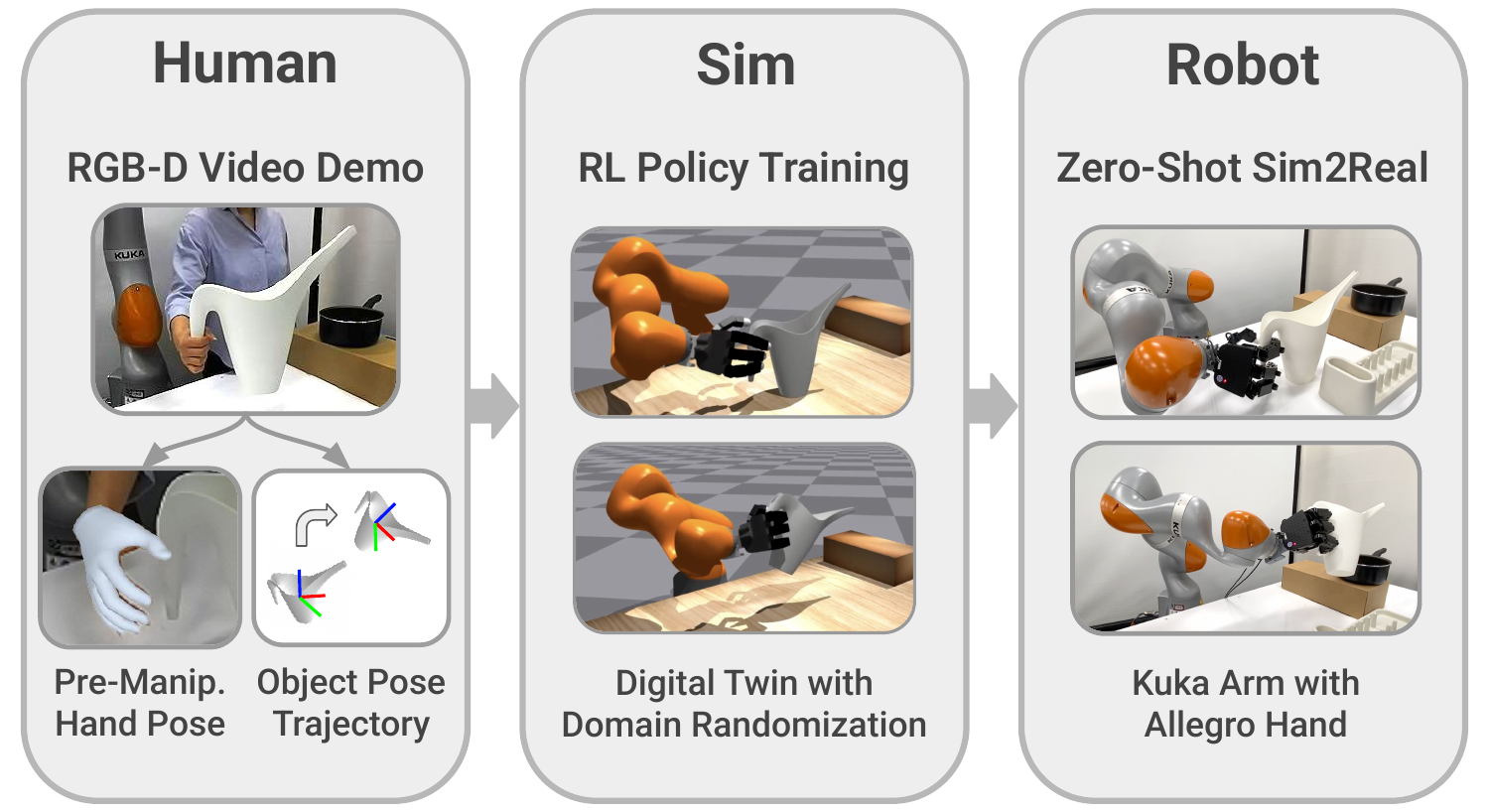}
\caption{\textbf{Our Framework.} \textsc{Human2Sim2Robot} learns dexterous manipulation policies from one human RGB-D video using object pose trajectories and pre-manipulation poses. These policies are trained with RL in simulation and transfer zero-shot to a real robot.}
\label{fig:splash}
\vspace{-0.4cm}
\end{wrapfigure}

Human-like dexterous hands have the potential to significantly advance robotic manipulation~\cite{qin2021dexmv,guzey2024hudor,chenobject}. However, the complexity of dexterous robot hands introduces substantial challenges for many existing robot learning methods. For example, imitation learning (IL) from human demonstration has shown success using simpler end-effectors with large amounts of training data~\cite{rt22023arxiv, rt12022, open_x_embodiment_rt_x_2023, jang2021bc, kim24openvla}, but collecting high-quality demonstrations for dexterous hands is far more difficult. Capturing high-quality 3D human hand motion typically relies on wearable sensors and teleoperation systems~\cite{wang2024dexcap,chen2024arcap}, which are expensive and difficult to scale. 

In contrast, videos of humans interacting with objects using their own hands are inexpensive to collect and offer a scalable alternative to traditional demonstration collection. However, leveraging them directly for robotic IL is challenging as they lack explicit robot action labels~\cite{ye2024latentactionpretrainingvideos}. One common approach to address this is by obtaining per-timestep human hand pose estimates and converting them to robot action labels via fingertip retargeting and inverse kinematics (IK)~\cite{wang2024dexcap,chen2024arcap}. However, this approach is often unreliable as hand pose reconstruction methods~\cite{pavlakos2024reconstructing} are susceptible to occlusion and sensor noise. Even with perfect pose estimates, this simple retargeting strategy often results in suboptimal robot trajectories due to morphological differences between the robot and human. These challenges are particularly punishing for contact-rich, dexterous manipulation~\cite{chen2025dexforceextractingforceinformedactions, ma2011dexterity, noll2024dexmanip}.

Existing IL methods are ill-equipped to address this issue as they rely on accurate correspondences between demonstrated and learned behaviors. Reinforcement learning (RL) offers a promising alternative to overcome these limitations by enabling robots to directly learn manipulation tasks using their own embodiment. However, RL has several limitations such as tedious, task-specific reward engineering and unfavorable sample complexity, making real-world policy training infeasible~\cite{torne2024rialto, zhu2020ingredientsrealworldroboticreinforcement}.

In this paper, we propose \textsc{Human2Sim2Robot}, a real-to-sim-to-real RL framework that addresses the limitations of existing methods and combines the best of both worlds: it only requires a single human RGB-D video demonstration and does not require any task-specific reward engineering. Crucially, we found that high-fidelity 3D human motion data is not necessary to learn robust dexterous manipulation policies. Instead, training an RL dexterous manipulation policy only requires two task-specific components that can be reliably extracted from the human video: (1) the object 6D pose trajectory, and (2) a single pre-manipulation hand pose. 

We use (1) to define an embodiment-agnostic, object-centric reward that specifies the desired task, and (2) to provide advantageous initialization for RL training and facilitate more efficient exploration. Formalizing an expert demonstration with these two components facilitates RL policy training with no task-specific reward tuning. Instead of directly learning state-action mappings, we use the demonstration for \textit{task specification} and \textit{guidance}, encouraging human-like behavior while allowing deviations when the human strategy is unsuitable for the robot's embodiment. This enables \textsc{Human2Sim2Robot} policies to achieve zero-shot sim-to-real transfer on a real-world dexterous robot, without requiring wearables, teleoperation, or large-scale data collection. 

To the best of our knowledge, \textsc{Human2Sim2Robot} is the first system that learns a robust real-world dexterous manipulation policy from only one human RGB-D video demonstration, bridging the human-robot embodiment gap across grasping, non-prehensile manipulation, and complex multi-step tasks. We achieve this with just \textit{a few minutes} of human effort end-to-end, from demonstration collection to digital twin construction. Our extensive ablation studies demonstrate the importance of our system's design decisions; while individual components have precedents, these works are often limited to simulation~\cite{qin2021dexmv,singh2024handobjectinteractionpretrainingvideos}, are not reactive closed-loop policies~\cite{chenobject,ye2023learning,kerr2024rsrd}, require significantly more demonstrations~\cite{qin2021dexmv,singh2024handobjectinteractionpretrainingvideos,kumar2022inverse,zakka2021xirl}, or only perform prehensile manipulation~\cite{qin2021dexmv,singh2024handobjectinteractionpretrainingvideos,ye2023learning,kerr2024rsrd}. 

\textsc{Human2Sim2Robot} policies can execute diverse real-world dexterous manipulation tasks, such as pouring from a pitcher, pivoting a box against a wall, and inserting a plate into a dishrack, without any task-specific reward tuning. In the single human demo regime, our method outperforms object-aware trajectory replay by $>$55\% and imitation learning by $>$68\% across all real-world tasks.


\section{Related Work}

\textbf{Visuomotor Imitation Learning for Robotics.} Visuomotor IL for robotic manipulation has shown success in learning from a large number of expert demonstrations~\cite{rt22023arxiv, rt12022, jang2021bc, kim24openvla,chi2024diffusion,zhao2024aloha} collected through teleoperation or specialized wearable equipment~\cite{wang2024dexcap,chi2024universal,lin2024datascalinglawsimitation}, which makes scaling data collection efforts expensive. In contrast, human videos are inexpensive and more intuitive to collect. Per-timestep human hand pose estimates can then be converted into robot action labels through IK-based retargeting~\cite{wang2024dexcap,chen2024arcap}. However, hand pose estimation noise and the human-robot embodiment gap often result in infeasible or suboptimal IK solutions for the robot embodiment, a challenge for visuomotor IL methods that directly rely on high-quality action labels. While human demonstrations provide useful guiding strategies for task completion, certain actions may not be suitable for robots given substantial embodiment differences. \textsc{Human2Sim2Robot} performs RL in simulation guided by a single human video demonstration. It encourages human-like behavior when beneficial while allowing deviations when the human strategy is unsuitable for the robot's embodiment.

\textbf{One-Shot Imitation Learning (OSIL).}
OSIL methods parallel our approach as a single demonstration is provided. Past work has performed object-aware retargeting to transfer the demonstrated trajectory to novel scenes~\cite{heppert2024ditto, vitiello2023one,okami2024}, leveraged object segmentation and visual servoing to adapt the single demonstration to a new scene~\cite{valassakis2022dome}, or augmented teleoperated demonstrations by retargeting and success filtering in a digital twin simulation~\cite{jiang2024dexmimicgen}. Though more data-efficient than visuomotor IL policies, OSIL methods suffer from limited generalization beyond the demonstrated actions. Simply replaying modifications of the single demonstration is unlikely to succeed in contact-rich settings requiring closed-loop, reactive behavior (e.g., variations in contact interactions or perturbations during policy rollout). Our insight is that using the human video to provide task specification and guidance for RL leverages this data source for robot learning  more effectively. This allows robots to develop effective strategies with their own embodiment, rather than rigidly imitating human behaviors.

\textbf{Reinforcement Learning for Robotics.} RL enables robots to learn complex behaviors through interaction with the environment. Real-world RL is often impractical due to slow training, safety concerns, manual environment resets, and difficult reward tuning~\cite{torne2024rialto, zhu2020ingredientsrealworldroboticreinforcement}. Sim-to-real RL circumvents these challenges and has led to breakthroughs in other robotic domains~\cite{cheng2023parkour, margolisyang2022rapid, miki2022quadruped,kaufmann2023champion,qi2022hand}. However, it remains underexplored for full arm-and-hand dexterous manipulation, as most prior work relies on simulation with non-physical, floating-hand models~\cite{chen2022towards, Rajeswaran-RSS-18, wan2023unidexgrasp++, xu2023unidexgrasp}. Among works that have demonstrated sim-to-real transfer, 
\citet{torne2024rialto} use demo-augmented RL, which requires many demonstrations collected with the same robot embodiment. \citet{chenobject} learn residual actions on top of an open-loop base trajectory learned from human data, but the resulting policy lacks the flexibility for error recovery and struggles under a large embodiment gap. In contrast, \textsc{Human2Sim2Robot} trains robust dexterous RL policies in simulation from minimal human input over a full arm-and-hand action space, which successfully transfer to the real world. Other works use inverse RL on human videos~\cite{kumar2022inverse, zakka2021xirl}, but inferring a reward function typically requires $\sim$100 demos. In contrast, our explicit object-centric reward works with a single demo. 

Prior works corroborate the observation that pre-grasp poses can accelerate policy learning and result in human-like grasps~\cite{Ciocarlie2007DexterousGV, dasari2023pgdm, luo2024graspingdiverseobjectssimulated}. However, these methods only focus on using pre-grasps from very similar embodiments for grasping tasks in simulation. We focus on training RL policies that transfer to the real world, overcome the human-robot embodiment gap, and perform prehensile and non-prehensile manipulation. These policies can be deployed zero-shot in real-world environments, as we build on recent work in sim-to-real transfer~\cite{torne2024rialto, lum2024dextrahg}. See Appendix~\ref{app:additional_related_work} for additional related work.

\begin{wrapfigure}{r}{0.55\textwidth}
\vspace{-1.7cm}
\centering
\includegraphics[width=\linewidth]{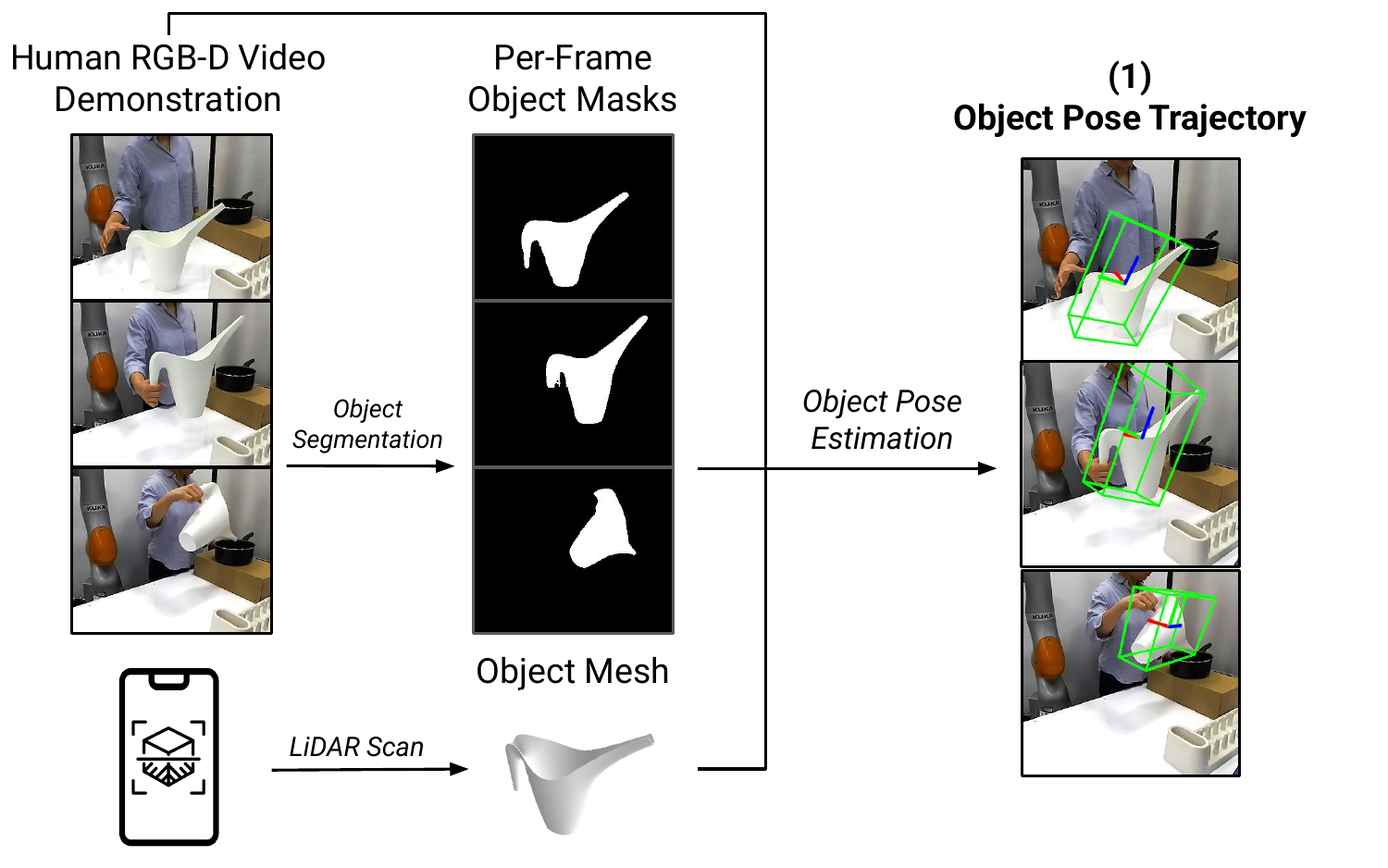}
\vspace{0.1cm}
\includegraphics[width=\linewidth]{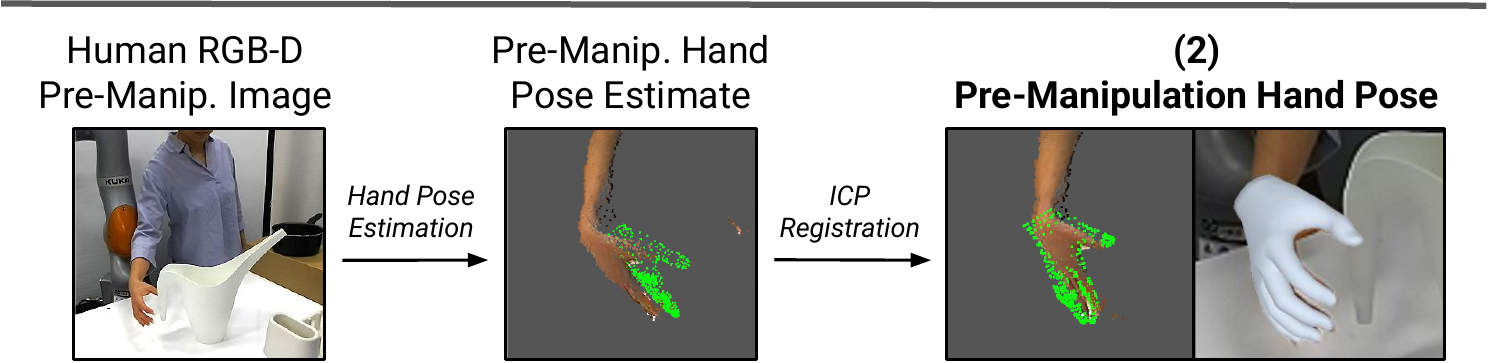}
\caption{\textbf{Human Demo Processing.} (1) The object pose trajectory defines an object-centric, embodiment-agnostic reward. (2) The pre-manipulation hand pose provides advantageous initialization for RL training.}
\label{fig:human_demo}
\vspace{-1.2cm}
\end{wrapfigure}

\section{Method}

We present \textsc{Human2Sim2Robot}, a real-to-sim-to-real RL framework for learning robust, dexterous manipulation policies from a single human-hand RGB-D video demonstration. Figure~\ref{fig:splash} shows an overview of our framework, and the following sections detail the key design decisions of our framework.

\subsection{Real-to-Sim \& Human Demo}
\label{sec:human_demo_real_to_sim}

We first create a digital twin of the robot's real-world environment and the target object to act as a policy training ground in simulation. The construction of the digital twin only takes $\sim$10 minutes of human effort (see Appendix~\ref{app:hamer_alignment} for a detailed time breakdown). We use off-the-shelf apps~\cite{KiriEngine2024,3DScannerApp2024} to capture a high-fidelity object mesh $\mathcal{O}$ and scene mesh $\mathcal{S}$.

Next, we record a single monocular RGB-D video demonstration $\{\bm{I}_t\}_{t=1}^T$ using a camera with known intrinsics and extrinsics, where each frame $\bm{I}_t \in \mathbb{R}^{H \times W \times 4}$ contains RGB-D data and $T$ is the total number of timesteps. Figure~\ref{fig:human_demo} visualizes how we process the human demonstration to obtain (1) an object pose target trajectory $\{\bm{T}^\text{target}_t\}_{t=1}^T$, and (2) a human hand pose trajectory represented as MANO~\cite{MANO:SIGGRAPHASIA:2017} parameters $ \left\{(\bm{\theta}_t, \bm{\beta}_t)\right\}_{t=1}^T
$. At each timestep $t$, $\bm{T}^\text{target}_t\in\text{SE}(3)$ is the object pose from the demonstration's object trajectory, $\bm{\theta}_t\in\mathbb{R}^{48}$ is the MANO hand pose parameter, and $\bm{\beta}_t\in\mathbb{R}^{10}$ is the MANO hand shape parameter. In our system, we extract the object pose trajectory using FoundationPose~\cite{wen2024foundationpose}, an open-source object pose detection model, which requires the scanned object mesh $\mathcal{O}$ and per-timestep object masks generated using Segment Anything Model 2 (SAM 2)~\cite{ravi2024sam2}, an open-source segmentation model. We extract per-timestep human hand poses using HaMeR~\cite{pavlakos2024reconstructing}, an open-source hand pose detection model. Since HaMeR takes RGB images as input, we use depth values from depth images and perform ICP registration to align the hand point clouds for obtaining accurate hand poses (see Appendix~\ref{app:hamer_alignment} for details on aligning HaMeR predictions with depth images).

We then determine the pre-manipulation hand pose at timestep $\tau = t_0 - t_{\text{offset}}$, where $t_0$ is the first timestep in which the object’s velocity exceeds a threshold $v_{\text{min}} = 5cm/s$, and $t_{\text{offset}}$ represents a fixed number of timesteps prior to the object's motion. We use $\bm{\theta}_{\tau}$ and $\bm{\beta}_{\tau}$ to compute the fingertip positions and the middle finger base knuckle pose as the human pre-manipulation hand pose. 

\begin{wrapfigure}{r}{0.55\textwidth}
\vspace{-0.4cm}
\centering
\includegraphics[width=\linewidth]{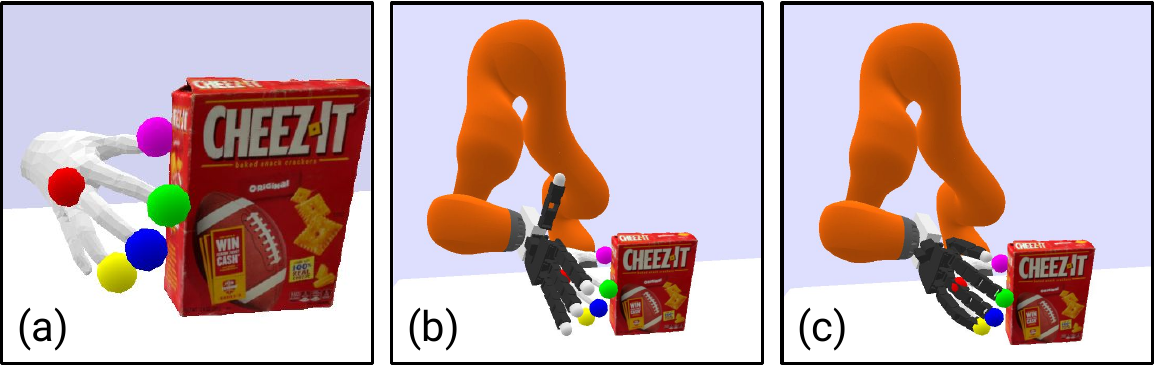}
\caption{\textbf{Human to Robot Hand Retargeting.} (a) Estimated MANO hand pose. Middle knuckle: red. Fingertips: pink, green, blue, yellow. (b) IK Step 1 (Arm): Align middle knuckle. (c) IK Step 2 (Hand): Align fingertips.}
\label{fig:retargeting}
\vspace{-0.4cm}
\end{wrapfigure}

Lastly, we retarget the human pre-manipulation hand pose to a robot hand pose through a two-step IK procedure using cuRobo~\cite{curobo_report23}. Figure~\ref{fig:retargeting} visualizes how we perform human-to-robot hand retargeting. In the first step, the robot arm's joint angles are adjusted to align the base position and orientation of the robot hand's middle knuckle with that of the human hand (with a small offset, see Appendix~\ref{app:retarget_details} for details). In the second step, the robot hand's joint angles are adjusted to align the robot's fingertip positions with the corresponding human fingertip positions. This generates a pre-manipulation robot hand pose $(\bm{T}^{\text{wrist}}, \bm{q}^{\text{hand}})$ for the object pose represented as $\bm{T}^{\text{target}}_\tau$, where $\bm{T}^{\text{wrist}}\in\text{SE}(3)$ is the robot wrist pose, and $\bm{q}^{\text{hand}}\in\mathbb{R}^{N^{\text{hand-joints}}}$ is the robot hand joint configuration. This IK process faithfully retargets the human pre-manipulation hand pose, while maintaining kinematic feasibility and alignment between the human and robot hand. 

The object pose trajectory provides \textit{task specification} by defining an object-centric trajectory-tracking reward for policy training. The pre-manipulation pose offers \textit{task guidance} by defining a good state initialization for exploration~\cite{dasari2023pgdm}. The pre-manipulation pose retargeting does not need to be extremely precise, as it is just a prior to facilitate exploration. Together, these abstractions guide RL policy training in simulation via reward guidance and advantageous state initializations.

\subsection{Simulation-based Policy Learning} \label{sec:sim_policy}

We create a training environment in the IsaacGym simulator~\cite{makoviychuk2021isaacgym} that matches the real-world environment, consisting of the robot, scene mesh $\mathcal{S}$, and object mesh $\mathcal{O}$, which takes only $\sim$10 minutes of human effort (see Section~\ref{sec:human_demo_real_to_sim}). We then train a policy using Proximal Policy Optimization~\cite{PPO} which outputs robot actions that move the object along the target trajectory, guided by the provided pre-manipulation pose. We emphasize that we primarily care about \textit{how the object moves}, rather than imitating the actions of the human demonstrator; the pre-manipulation pose serves as a rough prior, but the policy will learn to use the robot embodiment to achieve the desired object motion.

The reward given at timestep $t$ is an object-tracking reward, $r_t = r^{\text{obj}}_t$ defined as
\begin{equation} \label{eqn:r_obj}
    r^{\text{obj}}_t = \exp\bigl(-\alpha\,d(\bm{T}^{\text{target}}_{\tau+t}, \bm{T}^{\text{obj}}_t)\bigr),\quad \text{where} \quad d(\bm{T}_1, \bm{T}_2) = \sum_{i=1}^{N^{\text{anchor}}} \bigl\lVert\,\bm{T}_1\bm{k}_i - \bm{T}_2\bm{k}_i\bigr\rVert,
\end{equation}

\begin{wrapfigure}{r}{0.55\textwidth}
\centering
\includegraphics[width=\linewidth]{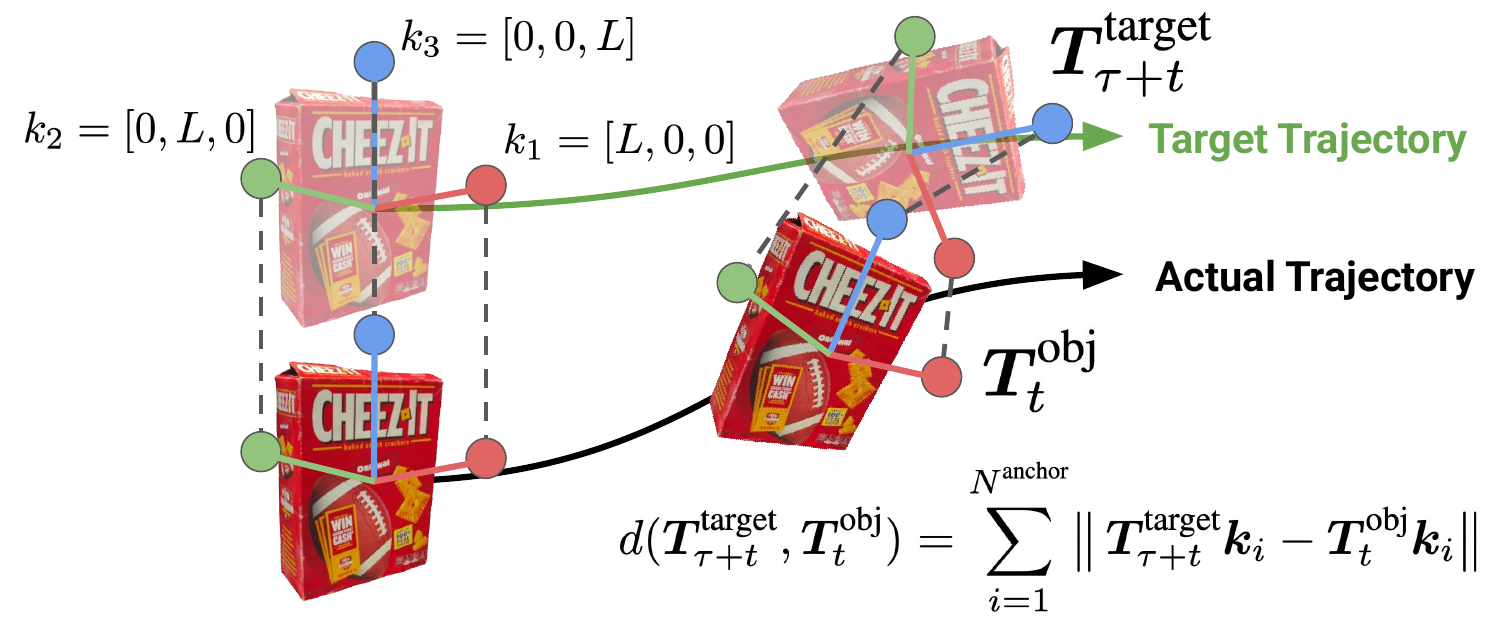}
\caption{\textbf{Object Pose Tracking Reward.} The agent is rewarded for minimizing distance between the current pose and target object pose $d(\bm{T}^{\text{target}}_{\tau+t}, \bm{T}^{\text{obj}}_t)$ using anchor points ${\bm{k}_i}$.}
\label{fig:object_pose_tracking_reward}
\vspace{-0.2cm}
\end{wrapfigure}

where $d$ is the relative pose distance function and $\alpha=10$. For most objects, we select $N^{\text{anchor}} = 3$, with $\bm{k}_1 = [L,0,0]$, $\bm{k}_2=[0,L,0]$, and $\bm{k}_3=[0,0,L]$ in the local object frame, where $L=0.2m$ is a distance parameter for orientation.
Figure~\ref{fig:object_pose_tracking_reward} visualizes the object pose tracking reward. This formulation integrates position and orientation naturally: larger $L$ emphasizes orientation by placing anchor points farther from the object’s origin. See Appendix~\ref{app:modifying_anchor_points} for reward function details.

The observation at timestep $t$ is $\bm{o}_t = [\bm{q}_t, \bm{\dot{q}}_t, \bm{X}^{\text{fingertips}}_t, \bm{x}^{\text{palm}}_t, \bm{X}^{\text{obj}}_t, \bm{X}^{\text{target}}_{\tau+t}]$, where $\bm{q}_t, \bm{\dot{q}}_t \in \mathbb{R}^{N^{\text{joints}}}$ are the robot's joint angles and velocities, $\bm{X}^{\text{fingertips}}_t \in \mathbb{R}^{N^{\text{fingers}} \times 3}$ are the fingertip positions, $\bm{x}^{\text{palm}}_t \in \mathbb{R}^3$ is the palm position, and $\bm{X}^{\text{obj}}_t, \bm{X}^{\text{target}}_{\tau+t} \in \mathbb{R}^{N^{\text{anchor}} \times 3}$ are the anchor point positions for the object and target poses. Here, $N^{\text{joints}}, N^{\text{fingers}} \in \mathbb{N}$ denote the number of robot joints and fingers, respectively.

The action at timestep $t$ is $\bm{a}_t =[\bm{x}^{\text{palm-target}}_t, \bm{r}^{\text{palm-target}}_t, \bm{x}^{\text{pca-target}}_t]$, where $\bm{x}^{\text{palm-target}}_t \in \mathbb{R}^3$ is the target palm center position, $\bm{r}^{\text{palm-target}}_t \in \mathbb{R}^3$ is the target palm orientation expressed as Euler angles, and $\bm{x}^{\text{pca-target}}_t \in \mathbb{R}^{N_{\text{pca}}}$ is the vector of target PCA values used to control the hand joints, where $N_{\text{pca}} = 5$ is the number of principal components. We use a geometric fabric controller and PCA-based hand action space following~\citet{lum2024dextrahg}, enabling human-like hand motions (see~\cite{lum2024dextrahg} for details).

We use an initial state distribution, guided by the human pre-manipulation hand pose, to simplify exploration and bias the policy toward human-like behavior. We construct this distribution by sampling the object pose around the trajectory's initial pose, then set the robot configuration to match the pre-manipulation hand pose with slight perturbation (see Appendix~\ref{app:init_distribution} for details). Given this perturbed pose, we compute a feasible arm joint configuration with IK~\footnote{We use cuRobo~\cite{curobo_report23} to perform parallelized, collision-free IK to ensure RL training is not bottlenecked by IK computations}. The environment is reset if the object is too far from the current target $d(\bm{T}^{\text{target}}_{\tau+t}, \bm{T}^{\text{obj}}_t) > D_{\text{max}}$, the robot palm is too far from the object $||\bm{x}^{\text{palm}}_t - \bm{x}^{\text{obj}}_t|| > D_{\text{max}}$, or the target trajectory is complete $\tau + t > T$, where $D_{\text{max}} = 0.25m$.

Real-world dynamics parameters (e.g., object mass, inertia, friction) are often unknown. To handle this uncertainty, we use domain randomization during simulation, enabling the policy to adapt to diverse dynamics and transfer to the real world. We train an LSTM-based policy that leverages a history of observations to handle this partial observability and noisy data. We train the policy using object poses instead of images to accelerate training and enhance robustness to visual variation. To improve resilience to pose errors and calibration noise, we add noise to object pose observations. We also apply random object forces to increase robustness to unexpected contacts, disturbances, and dynamics variation, enabling zero-shot sim-to-real transfer. See Appendix~\ref{app:sim_details} for training details.

After training, we deploy the policy on a real robot without additional fine-tuning (i.e., zero-shot). See Appendix~\ref{app:real_time} for sim-to-real inference-time details.


\section{Experiments \& Results} 

Our experiments aim to answer the following questions: 
(1) \textbf{Importance of Embodiment-Specific RL}: Do RL policies trained via \textsc{Human2Sim2Robot} outperform baselines on dexterous manipulation tasks? 
(2) \textbf{Importance of Object Pose Trajectory}: How effective is the object pose trajectory from a human demonstration as a dense reward for RL policy training, compared to other reward formulations? 
(3) \textbf{Importance of Pre-Manipulation Pose Initialization}: Does a pre-manipulation hand pose from a human demonstration provide more effective initialization for learning manipulation skills than generic initializations? 
(4) \textbf{Sufficiency of Pre-Manipulation Hand Pose}: How effective is a single pre-manipulation pose in guiding RL policy training, compared to alternatives that require full human hand trajectories? We evaluate \textsc{Human2Sim2Robot} in simulation and on a real robot across a diverse set of tasks and objects to answer these questions.

\subsection{Experimental Setup}

\textbf{Hardware Setup.} Our robot consists of a 16-DoF dexterous Allegro hand mounted on a 7-DoF KUKA LBR iiwa 14 arm. We used a ZED 1 stereo camera mounted on the table for recording both the human demonstration and real-time object pose estimation for policy input at test time. Our experiments are conducted in a tabletop setting with three static objects: a box, a large saucepan placed atop the box, and a dishrack. The tabletop and its static objects are captured in scene scan $\mathcal{S}$. Figure~\ref{fig:hardware_and_digital_twin} (Appendix) shows our hardware setup and objects, as well as the digital twin. 

\begin{wrapfigure}{r}{0.55\textwidth}
\vspace{-0.4cm}
\centering
\includegraphics[width=\linewidth]{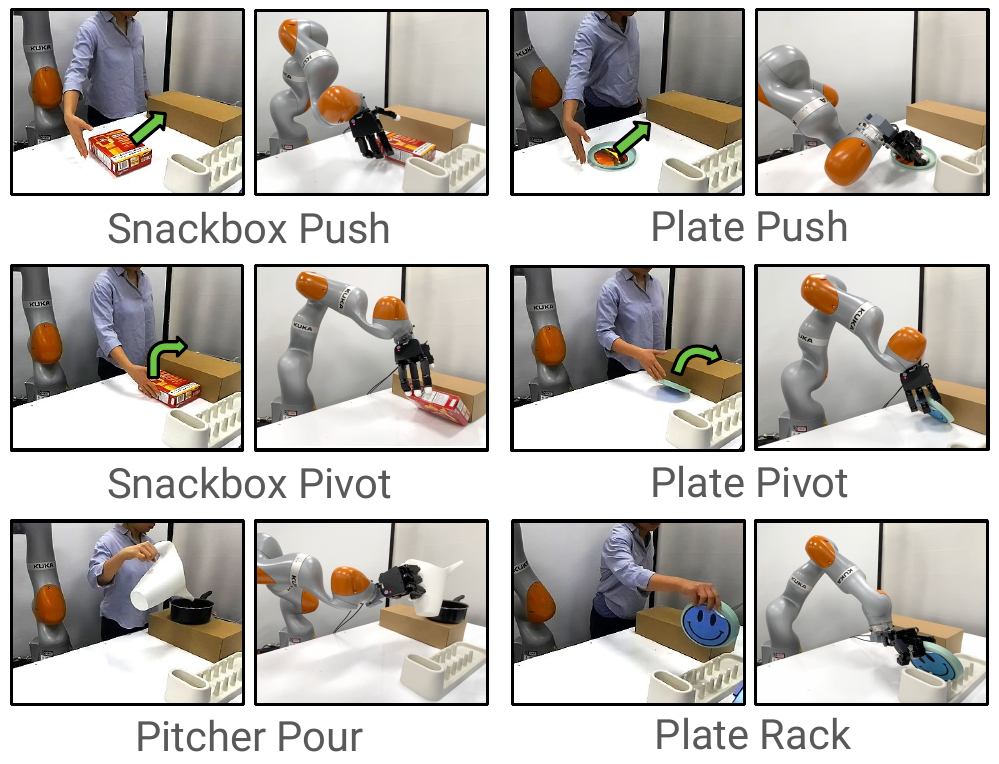}
\caption{\textbf{Task Visualization.} Our real-world tasks span grasping, non-prehensile manipulation, and extrinsic manipulation, with both the human demonstration and the resulting robot behavior. We also include three multi-step tasks that compose multiple skill types.}
\label{fig:task_visualization}
\vspace{-0.4cm}
\end{wrapfigure}

\textbf{Tasks \& Objects.} We perform experiments with three objects: \texttt{snackbox}, \texttt{pitcher}, and \texttt{plate}. Figure~\ref{fig:task_visualization} visualizes our tasks using these objects, which include grasping, non-prehensile manipulation, and extrinsic manipulation. We also explore multi-step tasks that compose sequences of these skills, such as pivoting the plate, lifting it, and placing it in a dishrack (see Appendix~\ref{app:tasks} for details).

\textbf{Simulation Ablation Setup.} For our ablation experiments, we evaluate all methods in simulation on the \texttt{plate-pivot-lift-rack} task, which is our most complex multi-step task. We train policies with three random seeds for all simulation results, and we compare them on their speed and stability of learning, their final achieved reward, and their qualitative behavior.

\subsection{Importance of Embodiment-Specific RL}

\begin{wrapfigure}{r}{0.65\textwidth}
\vspace{-0.6cm}
\centering
\includegraphics[width=\linewidth]{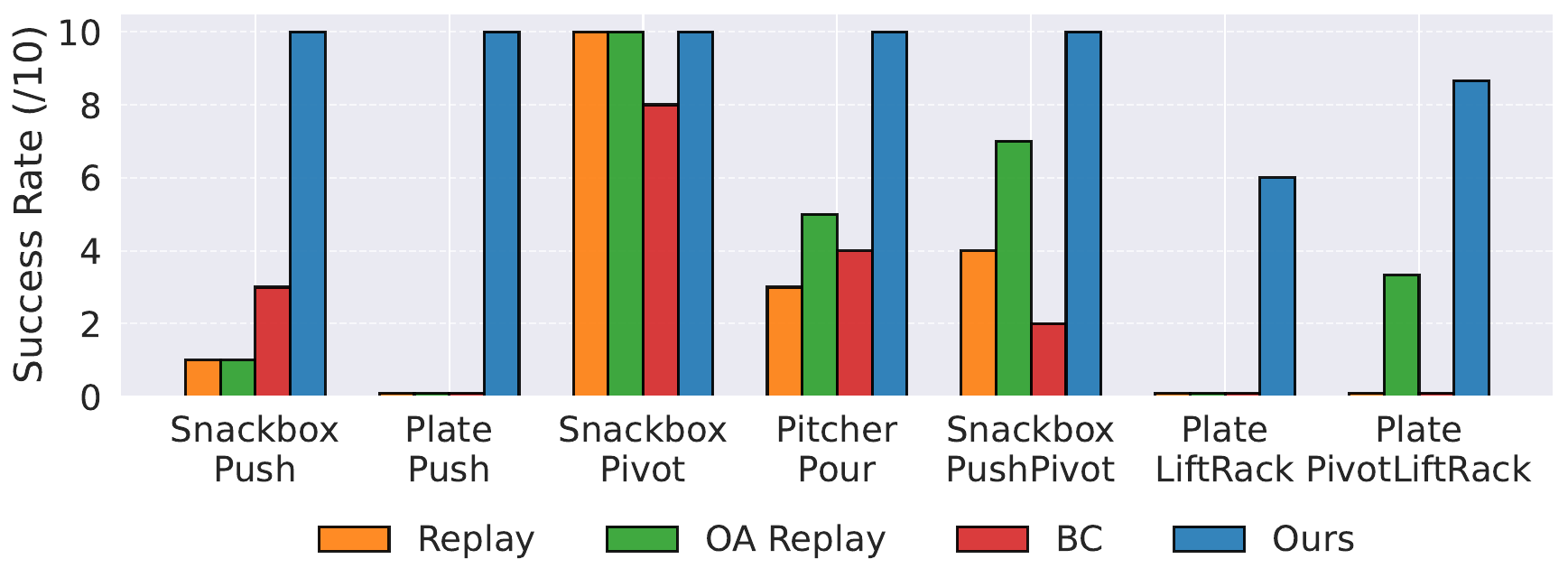}
\caption{\textbf{Real-World Success Rates.} \textsc{Human2Sim2Robot} policies outperform Replay by 67\%, Object-Aware (OA) Replay by 55\%, and Behavior Cloning (BC) by 68\% across all tasks.}
\label{fig:success_rates}
\vspace{-0.6cm}
\end{wrapfigure}

In real-world experiments, we compare \textsc{Human2Sim2Robot} policies to non-RL baselines to evaluate the impact of closed-loop, embodiment-specific RL. These baselines require robot action labels for the entire task, which are obtained by performing hand pose estimation and human-to-robot retargeting for \textit{every} frame of the video.

We evaluate against three baselines across seven real-world tasks (see Appendix~\ref{app:baseline_details} for baseline details):
(1) \textbf{Replay}: Replays the retargeted trajectory open-loop by setting PD targets to these positions;
(2) \textbf{Object-Aware (OA) Replay}: Like \textbf{Replay}, but warps the trajectory by the relative transform between initial object pose in the human demo and at test time (similar to~\cite{vitiello2023one,okami2024});
and (3) \textbf{Behavior Cloning (BC)}: Trains a closed-loop diffusion policy~\cite{chi2023diffusionpolicy} on 30 demos (same number as in \cite{guzey2024hudor}), generated from our one demo by sampling object poses (same range as our RL training), performing OA Replay, and using these trajectories as demo data.

For each task, we evaluate the success rate of the task-specific RL policy and the baselines across 10 policy rollouts (Figure~\ref{fig:success_rates}). In all tasks, \textsc{Human2Sim2Robot} policies substantially outperform the baselines. \textbf{Replay} was unsuccessful on most tasks, but performed well on tasks requiring low precision like \texttt{snackbox-pivot}. \textbf{OA Replay} performed better than \textbf{Replay} as it accounts for randomizations in initial pose, but still had many failures due to (a) hand pose estimation errors, (b) morphological differences between the robot and human, and (c) non-reactive open-loop control. \textbf{BC} performed similarly to \textbf{Replay}, which can be attributed to the low-quality dataset (actions computed from noisy hand pose estimations) and compounding errors throughout policy rollouts. While retargeted robot demos can succeed on simpler tasks, they often fail on harder multi-step tasks. \textbf{Ours} does not simply imitate human behaviors, but adapts the behavior for the robot embodiment, resulting in much higher success rates across all tasks (see Appendix~\ref{app:qualitative_analysis} for further analysis).

\subsection{Importance of the Object Pose Trajectory}

We run ablation experiments in simulation to study the importance of the object pose tracking reward, comparing against: (1) \textbf{Fixed Target}: In $r^{\text{obj}}_t$ (Eq.~\ref{eqn:r_obj}), we replace the current target object pose $\bm{T}^{\text{target}}_{\tau+t}$ with a final one $\bm{T}^{\text{target}}_T$; (2) \textbf{Interpolated Target}: In $r^{\text{obj}}_t$ (Eq.~\ref{eqn:r_obj}), we replace the current target pose $\bm{T}^{\text{target}}_{\tau+t}$ with an interpolated pose between the initial and final target pose: $\textsc{interp}\left(\bm{T}^{\text{target}}_{\tau}, \bm{T}^{\text{target}}_T, t/(T-\tau)\right)$, where $\textsc{interp}: \text{SE}(3) \times \text{SE}(3) \times [0, 1] \to \text{SE}(3)$ linearly interpolates position and uses slerp for orientation; (3) \textbf{Downsampled Trajectory}: In $r^{\text{obj}}_t$ (Eq.~\ref{eqn:r_obj}), we replace the current target object pose $\bm{T}^{\text{target}}_{\tau+t}$ with the downsampled pose $\bm{T}^{\text{target}}_{\tau + t_\text{down}}$, where $t_{\text{down}} = \lfloor t / D \rfloor \cdot D$ and $D$ is the downsampling factor. This reduces the temporal resolution of the human demonstration trajectory into a series of key poses.

\begin{wrapfigure}{r}{0.55\textwidth}
\vspace{-0.4cm}
\centering
\includegraphics[width=\linewidth]{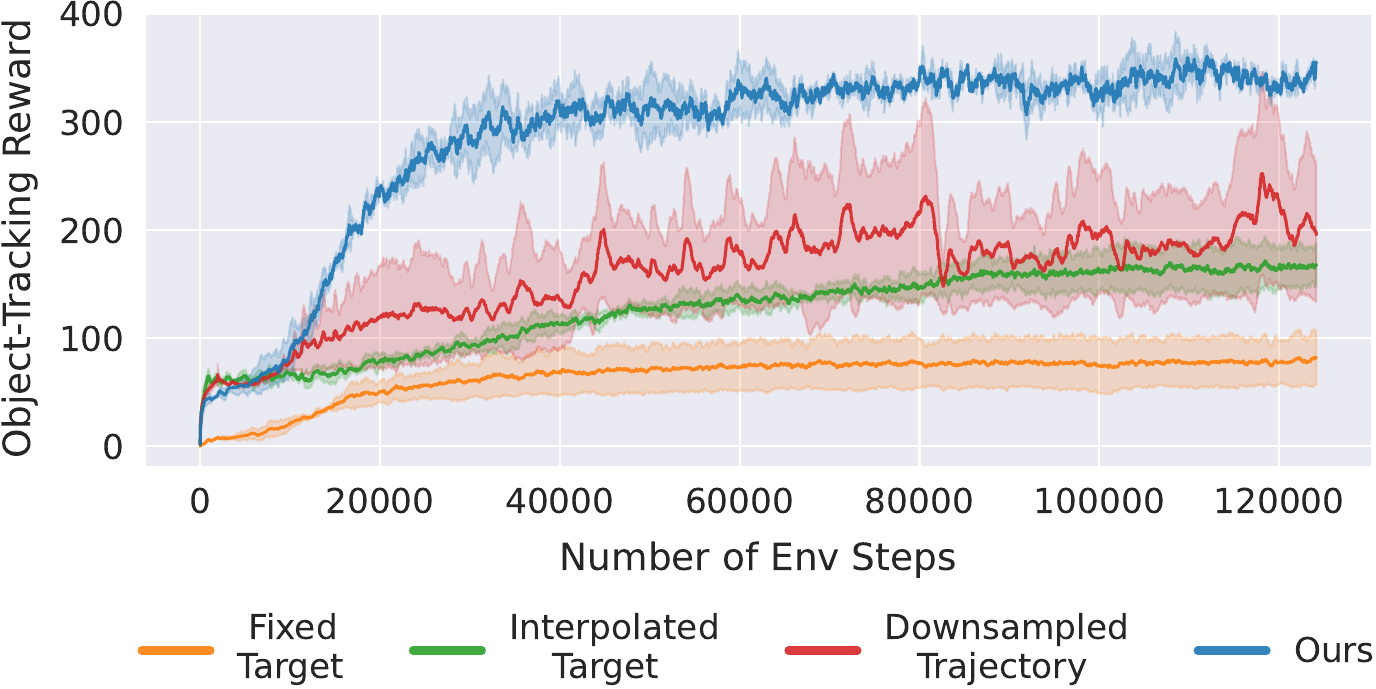}
\caption{\textbf{Object Pose Tracking Reward Ablation.} Reward curves comparing different object rewards.}
\label{fig:object_reward_ablation}
\vspace{-0.4cm}
\end{wrapfigure}

Figure~\ref{fig:object_reward_ablation} shows that \textsc{Human2Sim2Robot} achieves a substantially higher average reward than the other methods. The plate lying flat on the table is too large to be directly grasped, therefore the optimal strategy is to use extrinsic manipulation leveraging the wall to pivot the plate into a graspable position, as demonstrated in the human video. \textbf{Fixed Target} and \textbf{Interpolated Target} encourage the policy to greedily move the plate directly to the target in the air, but they struggle to pick up the plate and get stuck in a local minimum. The policy does not explore extrinsic manipulation because this requires navigating to low reward regions for a long period. \textbf{Downsampled Trajectory} is able to learn the task, but it takes longer to learn due to the weaker learning signal. It produces jerky motions in hopes of maximizing reward by tracking the waypoints that jump suddenly. \textbf{Ours} uses the full dense object pose trajectory, and it achieves the strongest performance and converges the fastest. This underscores the effectiveness of our dense, object-centric, and embodiment-agnostic reward function.

\subsection{Importance of Pre-Manipulation Pose Initialization}

We run ablation experiments in simulation to study the importance of pre-manipulation pose initialization, comparing against: (1) \textbf{Default Initialization}: We initialize the robot configuration at a default rest pose that is not close to the object; (2) \textbf{Overhead Initialization}: We compute a joint configuration with IK setting the robot palm 5cm above the object. We set the hand joint angles to a default open hand pose; (3) \textbf{Pre-Manipulation Far}: We initialize the robot with the pre-manipulation hand pose, but adjust the arm joint angles to move the robot palm 20cm away from the object.

\newpage  
\begin{wrapfigure}{r}{0.55\textwidth}
\vspace{-0.4cm}
\centering
\includegraphics[width=\linewidth]{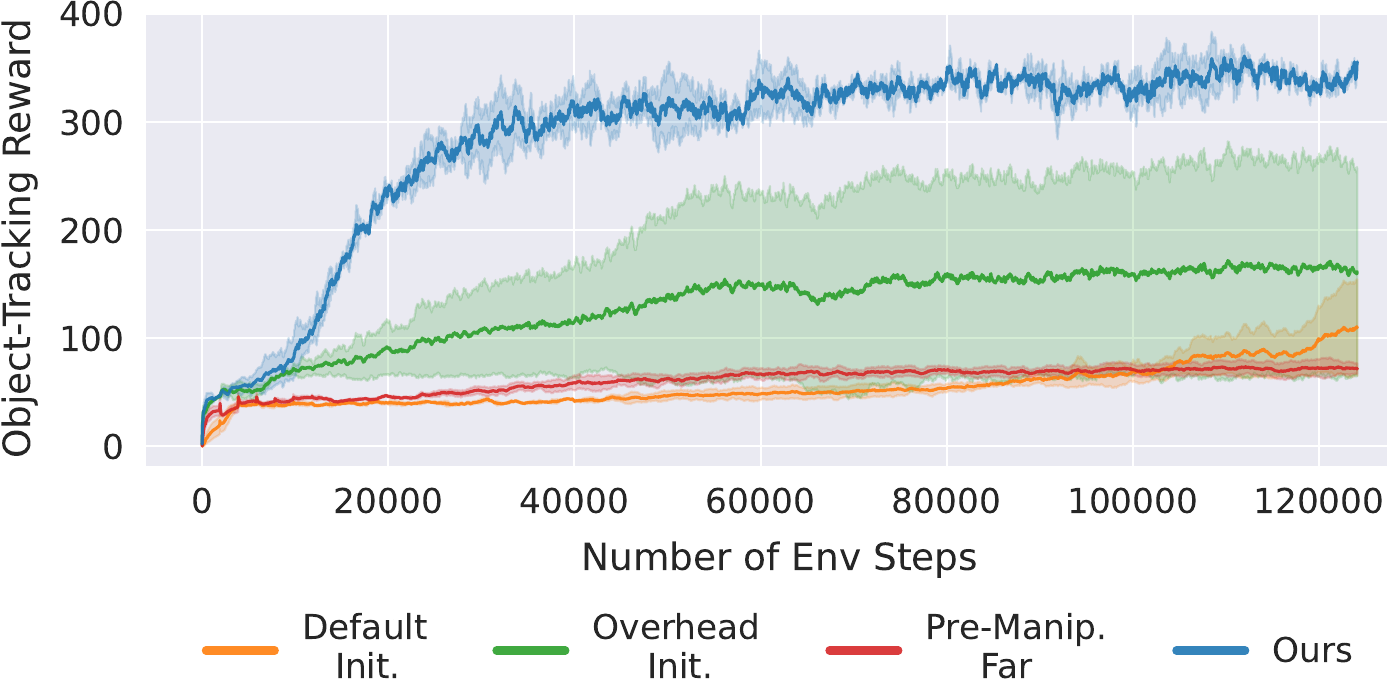}
\caption{\textbf{Pre‑Manip. Pose Ablations.} Reward curves comparing different initialization strategies.}
\label{fig:premanip_ablation}
\vspace{-0.4cm}
\end{wrapfigure}

Figure~\ref{fig:premanip_ablation} shows that \textsc{Human2Sim2Robot} achieves a substantially higher average reward than the other methods. \textbf{Default Initialization} and \textbf{Pre-Manipulation Far} perform the worst due to exploration challenges from starting with the hand too far away from the object. \textbf{Overhead Initialization} performs slightly better because it is initialized closer to the object, but fails to converge on a successful policy because the overhead grasp provides a disadvantageous prior, as it is not the optimal approach for performing the task. \textbf{Ours} achieves the highest reward by providing an advantageous initialization, which minimizes exploration challenges. See Appendix~\ref{app:qualitative_analysis} for additional qualitative analysis.

\subsection{Sufficiency of Single Pre-Manipulation Pose}

We run ablation experiments in simulation to compare pre-manipulation pose initialization to methods that require the full human hand trajectory: (1) \textbf{Hand Tracking Reward}: We add $r^{\text{hand}}_t = \exp\left(-\alpha \|\bm{X}^{\text{fingertips}} - \bm{X}^{\text{desired-fingertips}}\|\right)$ to encourage tracking the human hand trajectory. The total reward is $r_t = r^{\text{obj}}_t + r^{\text{hand}}_t$; (2) \textbf{Residual Policy}: Using the same object-centric reward, we replay the retargeted robot trajectory open-loop while learning delta PD joint targets (similar to~\cite{chenobject}).

\begin{wrapfigure}{r}{0.55\textwidth}
\vspace{-0.4cm}
\centering
\includegraphics[width=\linewidth]{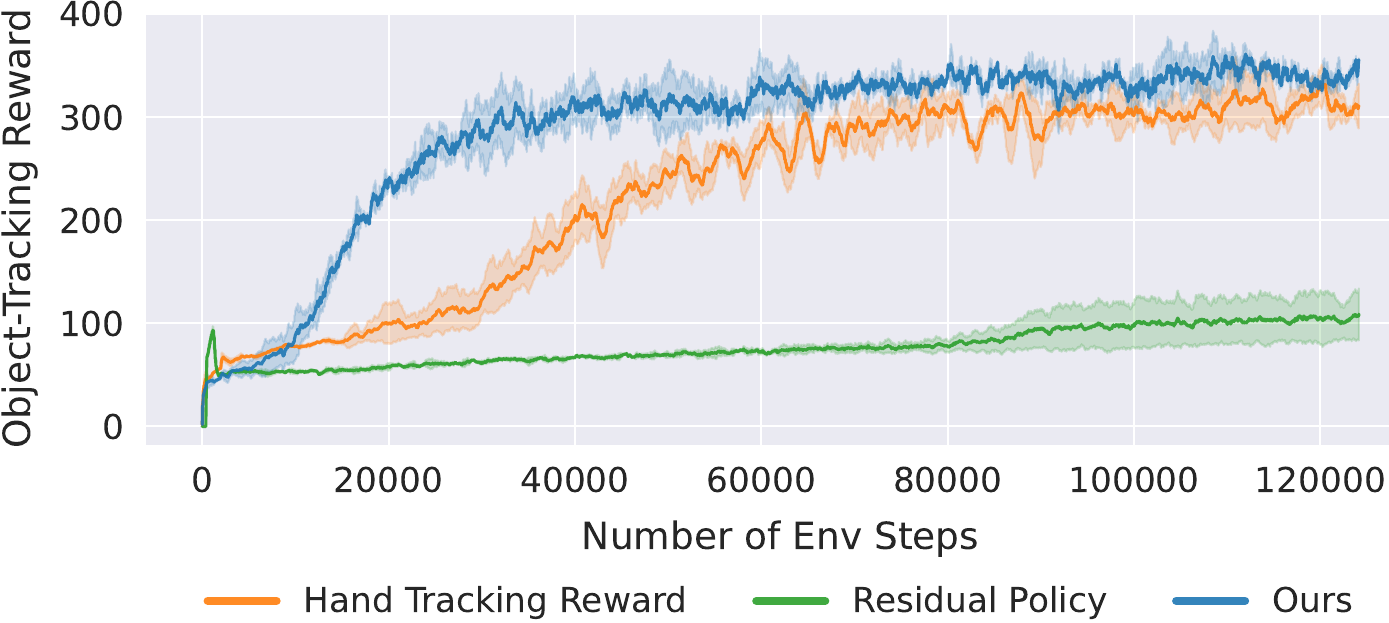}
\caption{\textbf{Full Hand Trajectory Ablation.} Reward curves comparing our method with methods that require the full human hand trajectory.}
\label{fig:hand_tracking_ablation}
\vspace{-0.4cm}
\end{wrapfigure}

Figure~\ref{fig:hand_tracking_ablation} shows the reward curves of these methods. \textbf{Hand Tracking Reward} is able to learn an effective policy, but it learns more slowly because it initially focuses on improving its hand-tracking reward, which is not always conducive to policy performance. When trained to convergence, \textbf{Hand Tracking Reward} does not show any substantial improvement over our method, despite requiring additional data and supervision. \textbf{Residual Policy}'s performance is substantially worse because inaccurate hand pose estimation results in a poor base motion that is difficult to learn an effective residual policy for. In contrast, \textbf{Ours} is able to effectively learn the task without requiring the full human hand trajectory as additional supervision.


\section{Conclusion}  

We present \textsc{Human2Sim2Robot}, a real‑to‑sim‑to‑real RL framework for learning robust dexterous manipulation policies from a single human hand RGB-D video demonstration. Our method facilitates training RL policies in simulation for dexterous manipulation by formalizing tasks through object‑centric rewards and a pre‑manipulation hand pose. \textsc{Human2Sim2Robot} addresses several key challenges in general dexterous manipulation including efficient exploration, eliminating reward engineering effort, bridging the human-robot embodiment gap, and robust sim-to-real transfer. Our policies show significant improvements over existing methods across grasping, non-prehensile manipulation, and extrinsic manipulation tasks. Overall, \textsc{Human2Sim2Robot} represents a step forward in facilitating scalable and robust training of real-world dexterous manipulation policies.

\section{Limitations \& Future Work} 

In this paper, we evaluate \textsc{Human2Sim2Robot}'s ability to bridge the human-robot embodiment gap using a Kuka arm and Allegro hand. While we have not conducted extensive quantitative evaluations on other embodiments, we expect our framework to be similarly effective for other embodiments, as our framework was deliberately designed to be \textit{embodiment-agnostic}. Specifically, our object-centric reward function and training pipeline do not rely on embodiment-specific assumptions. As a preliminary test for this design choice, we present initial experiments on both the LEAP Hand~\cite{shaw2023leaphand} and UMI gripper~\cite{chi2024universal}, which achieved strong performance with minimal modifications (see Appendix~\ref{app:other_embodiments} for details). We believe these early results are promising, and future work can further validate our approach's generality by applying it to a broader range of robot hands and arms.

In addition, \textsc{Human2Sim2Robot} focuses on tasks specified by the pose trajectory of a single object. Adapting the method to handle multiple objects is possible with our framework, but would require additional modifications. \textsc{Human2Sim2Robot} also assumes a high-quality object tracker (in our case, an object pose estimator) and simulator. Our tasks therefore only feature rigid-body objects and environments, which can be efficiently tracked with existing pose estimators and simulated using existing rigid-body simulators. Extending to articulated or deformable objects would necessitate adapting \textsc{Human2Sim2Robot} to  different state estimation and simulation approaches. 

While using 6D object pose as policy input enhances robustness against visual distractors and simplifies RL training and sim-to-real transfer, our pose estimator has limitations as it struggles with pose ambiguity of symmetric objects and sensor noise from reflective objects. Invariance to pose ambiguity of symmetric objects could be achieved by modifying the anchor points used to determine object-centric rewards, for instance by automatically determining the axis of symmetry given the object mesh~\cite{PEI19941193} and removing the points orthogonal to this axis of rotation (see Appendix~\ref{app:modifying_anchor_points}). For reflective objects, these challenges can be addressed by distilling the pose-based policies into image-based policies, similar to prior work~\cite{torne2024rialto,lum2024dextrahg}.   
Finally, \textsc{Human2Sim2Robot} currently trains robust single-object, single-task policies. In future work, we can explore training generalist multi-task, multi-object policies that are conditioned on object shape and desired object trajectory. This can be achieved using multi-task RL or teacher-student distillation. Future work can also investigate extensions to \textsc{Human2Sim2Robot} for bimanual manipulation or whole-body manipulation tasks.


\clearpage
\acknowledgments{We thank the reviewers for their helpful suggestions and feedback. This work is supported by Stanford Human-Centered Artificial Intelligence, the National Science Foundation under Grant Numbers 2153854 and 2342246, and the Natural Sciences and Engineering Research Council of Canada (NSERC) under Award Number 526541680.}


\bibliography{references}  

\clearpage

\appendix

{\Large \textbf{Appendix}}

\section{Real-to-Sim \& Human Demo Processing Details}
\label{app:hamer_alignment}

\subsection{Digital Twin Construction Details}

\begin{wrapfigure}{r}{0.55\textwidth}
\vspace{-0.4cm}
\centering
\includegraphics[width=\linewidth]{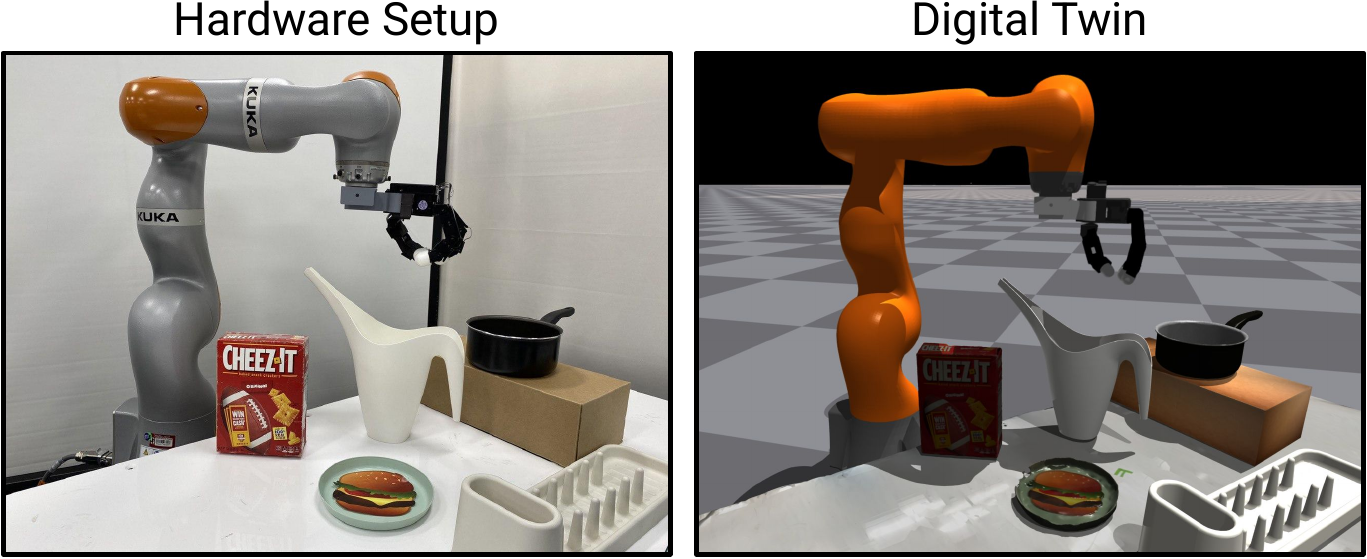}
\caption{\textbf{Hardware Setup \& Digital Twin.} Experiments are conducted in a tabletop setting with a static box, saucepan, and dishrack.}
\label{fig:hardware_and_digital_twin}
\vspace{-0.4cm}
\end{wrapfigure}

In this section, we describe the construction of the digital twin simulation environment (i.e., real-to-sim) and how we process the human demonstration. We start by taking a scan of the real-world environment for transfer to simulation. We used the off-the-shelf LiDAR scanning app called 3D Scanner App \cite{3DScannerApp2024} to perform a detailed scene scan of the workspace (including static objects on the tabletop) and the robot. We then used this scene scan to align the robot, tabletop, and static objects in our digital twin simulation. The scene scan took $\sim$3 minutes and alignment of the assets in simulation took another $\sim$2 minutes. This process was performed once and the same simulation environment was used for all tasks. Next, we need to take an object scan to obtain the object mesh used for object pose tracking. We use another off-the-shelf LiDAR scanning app called Kiri Engine \cite{KiriEngine2024} to do this, as we find it is better for scanning small objects than the 3D Scanner App (conversely, 3D Scanner App seems better for larger scene scans than Kiri Engine). This takes $\sim$2 minutes per object, and we only need to do this once per object used in our experiments. All relevant code for the real-to-sim pipeline can be found \href{https://github.com/tylerlum/human2sim2robot/tree/main/human2sim2robot/real_to_sim}{here}.

\subsection{Human Demo Processing Details}
For processing the human demonstration, each task takes 0.5 minutes to obtain an RGB-D video demonstration. The human demonstration is then processed in three steps: (i) generating the per-frame object and hand masks using SAM 2~\cite{ravi2024sam2}; (ii) generating the object pose trajectory by passing in the RGB-D video frames, object segmentation mask frames, and object mesh into FoundationPose \cite{wen2024foundationpose}; (iii) obtaining the pre-manipulation hand pose by passing the corresponding RGB frame into HaMeR~\cite{pavlakos2024reconstructing}, then performing depth alignment with the segmented hand depth values. In total, demonstration processing takes $\sim$5-10 minutes, with only a few seconds of human effort. All relevant code for processing human video demonstrations can be found \href{https://github.com/tylerlum/human2sim2robot/tree/main/human2sim2robot/human_demo}{here}.

\subsection{Depth Alignment of HaMeR Hand Pose Estimates}

HaMeR~\cite{pavlakos2024reconstructing} predicts MANO hand pose and shape parameters from a single RGB image. While its 2D projections are generally reliable, the absence of depth information leads to significant 3D errors, which can be as large as 20–30 cm, particularly when the hand is far from the camera. Such discrepancies result in suboptimal pre-manipulation hand poses, adversely affecting reinforcement learning (RL) initialization and methods that require a full human hand trajectory.

To address this issue, we incorporate depth information to refine HaMeR’s hand pose predictions. First, we obtain the initial MANO hand estimate from HaMeR. Next, we generate a hand segmentation mask using SAM 2~\cite{ravi2024sam2} and extract the corresponding 3D hand points from the depth image using the camera parameters. Next, we compute the 3D positions of the predicted MANO hand vertices visible to the camera and use a clustering algorithm to filter out erroneous depth points. Specifically, we construct a KD-tree from the extracted 3D points and identify pairs of points within 5 cm using nearest-neighbor queries. We then represent these points as a graph, where edges connect nearby points. We assume that the hand point cloud is the largest connected component, so we only retain the points in this connected graph, which mitigates the impact of depth noise and segmentation errors from the arm or background. Finally, we apply Iterative Closest Point (ICP) registration to align the MANO prediction with the segmented depth-based point cloud. This approach effectively improves alignment in most frames. However, accuracy degrades when significant occlusions occur, such as when fingers grasp around an object and face away from the depth camera.

\section{Human-to-Robot Retargeting Details}
\label{app:retarget_details}

To perform arm IK, we use the parallelized IK solver provided by cuRobo~\cite{curobo_report23}. The solver generates 100 unique solutions, each seeded with 20 joint configurations. These configurations consist of a selected joint configuration combined with random noise sampled from a normal distribution with a standard deviation of 15 degrees. From these solutions, we select the one closest to the selected joint configuration that meets the desired target within 5 cm for position and 3 degrees for orientation. The pose target is the position of middle knuckle (called ``middle\_0'') with a 3cm offset in the negative direction of the palm normal and a 3cm offset in the negative direction of the wrist to middle knuckle, and the orientation of the wrist (called ``global\_orient'') with a rotation offset accounting for the difference in orientation convention of the wrist and middle knuckle.

For hand IK, we use PyBullet's IK solver~\cite{coumans2019}. The Allegro hand is moved to the pose from the previous arm IK solution, after which we solve IK for each finger to reach the specified fingertip targets. A default hand pose is used as the rest pose. Since achieving precise alignment with all fingertip targets is not always possible, PyBullet's solver provides reasonable solutions in these cases, whereas cuRobo's solver may yield poor results when it fails. We use position targets at the human hand fingertips (called ``index\_3'', ``middle\_3'', ``ring\_3'', and ``thumb\_3'') with no adjustments.

To retarget the pre-manipulation hand pose, we execute the above process once, using the default upright arm configuration as the selected joint configuration for seeding the solver and selecting the best solution. During this step, we apply collision-free arm IK to ensure the robot avoids contact with the environment or objects, as this joint configuration should not be in collision.

For retargeting the full hand pose trajectory, we iteratively solve for each hand pose, using the previously computed solution as the selected joint configuration to seed the solver. This approach ensures consistency in joint angles between frames, resulting in smoother motion. While solving, we enforce collision-free arm IK to avoid contact with the environment. However, we do not enforce object collision avoidance, as contact with the object is often necessary and both hand and object pose estimates are imperfect. If the IK process fails for a particular pose, we skip that index and proceed to the next one, keeping track of the corresponding timesteps.

When evaluating solutions, we select the one closest to the selected joint configuration based on the infinity norm, $||\bm{q}_{\text{solution}} - \bm{q}_{\text{selected}}||_\infty$.

After retargeting the full hand pose trajectory, we modify the trajectory to ensure smoothness. At each timestep, we compute the arm joint velocity with first-order finite-differencing. If any arm joint velocity exceeds its velocity limit (indicating a discontinuity or sudden jump in joint positions within a short time period), we increase the time interval between those points and interpolate the intermediate values. This ensures that the joint velocity limits are respected, resulting in a smooth trajectory that can be safely executed on the robot.

\newpage 

\section{Reward Function Details}
\label{app:modifying_anchor_points}

\begin{wrapfigure}{r}{0.6\textwidth}
    \vspace{-0.4cm}
    \centering
    \includegraphics[width=\linewidth]{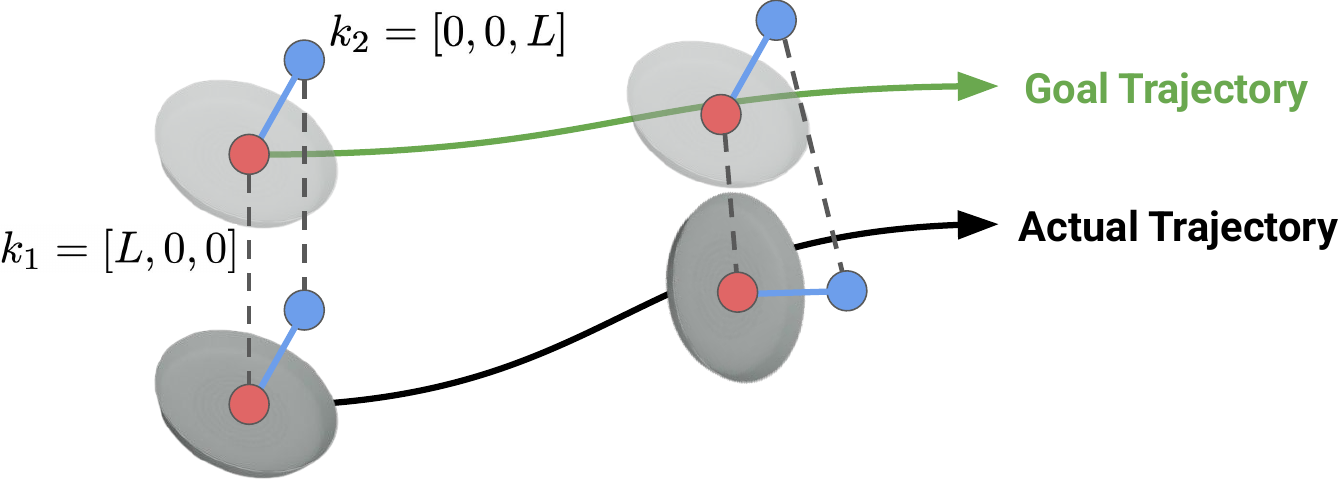}
    \caption{\textbf{Modifying Anchor Points.} For rotationally symmetric objects, we remove anchor points orthogonal to the axis of symmetry (automatically determined~\cite{PEI19941193}). This modified object pose representation is used for both reward computation and policy observation.}
    \label{fig:modifying_anchor_points}
    \vspace{-0.4cm}
\end{wrapfigure}

Our reward function formulation is flexible enough to handle rotational symmetries or axis invariances by removing or repositioning anchor points as needed. Apart from the adjustments for rotationally symmetric objects, we use the same value for hyperparameter $L$ for all tasks in the experiments. This highlights the generality and robustness of the proposed reward specification, making it broadly applicable across various tasks and objects. 

Figure~\ref{fig:modifying_anchor_points} shows how anchor points can be modified to accommodate rotational invariance for rotationally symmetric objects. Our use of anchor points is a very flexible representation to parameterize a pose tracking reward function.

\section{Initial State Distribution Sampling}
\label{app:init_distribution}
In this section, we describe sampling the initial state distribution for RL training in simulation. The default initial object pose $\bm{T}^{\text{target}}_{\tau}$ is the object pose from the human demonstration at the timestep $\tau$ at which the premanipulation hand pose was acquired. To sample the initial object pose $\bm{T}^{\text{obj}}_{1}$, we sample random pose noise $\bm{T}^{\text{random}}\in\text{SE}(3)$, then compute $\bm{T}^{\text{obj}}_{1} = \bm{T}^{\text{target}}_{\tau}\bm{T}^{\text{random}}$. The rotation component of $ \bm{T}^{\text{random}} $ is sampled as a yaw angle from 
$y \sim \mathcal{U}(-\theta_{\text{max}}, \theta_{\text{max}})$, where $ \theta_{\text{max}} = 20^\circ$, with roll and pitch fixed at zero, restricting rotation to the horizontal plane only. The translation component of $ \bm{T}^{\text{random}} $ is sampled as $x, y \sim \mathcal{U}(-t_{\text{max}}, t_{\text{max}})$, where $ t_{\text{max}} = 0.1m $, with $z = 0$, ensuring that objects remain at their original height. Next, we compute the relative transformation $\bm{T}^{\text{relative}} = \left(\bm{T}^{\text{target}}_{\tau}\right)^{-1}\bm{T}^{\text{wrist}}$ to determine the new pre-manipulation wrist pose as $\bm{T}^{\text{obj}}_{1}\bm{T}^{\text{relative}}$. An additional small amount of noise is added to the new wrist pose and hand joint angles. 

\section{Simulation Training Details}
\label{app:sim_details}

We train our policy using Isaac Gym~\cite{makoviychuk2021isaacgym}, a high-performance GPU-accelerated simulator that enables the parallel simulation of 4096 robots per GPU. Each policy is trained on a single NVIDIA A100 GPU with 40 GB of VRAM. This configuration allows us to achieve a simulation speed of approximately 7k frames per second (FPS). Each frame corresponds to a single action step with a control timestep of 66.7 ms (15 Hz), subdivided into 8 simulation timesteps of 8.33 ms (120 Hz).

Our training duration ranges from approximately 5 to 24 hours wall-clock time depending on the task, amounting to around 0.6 billion frames, which corresponds to roughly one year of simulated experience (0.6B / 15 / 3600 / 24 / 365).

We train our policies using Proximal Policy Optimization (PPO)~\cite{PPO} with rl-games~\cite{rl-games2021}, a highly optimized GPU-based implementation that employs vectorized observations and actions for efficient training. The policy is trained with learning rate $5 \times 10^{-4}$, discount factor $\gamma = 0.998$, entropy coefficient of 0, and PPO clipping parameter $\epsilon_{\text{clip}} = 0.2$. Additionally, we normalize observations, value estimates, and advantages, and train the policy using four mini-epochs per policy update.

We use $t_{\text{offset}} = 30$ (at 30Hz, 1 second) by default, though we slightly vary $t_{\text{offset}}$ depending on how quickly the human demonstration was performed. In our \textbf{Downsampled Trajectory} ablation experiment, we use $D = 90$ (at 30Hz, 3 seconds), which resulted in a sparse downsampled trajectory.

Although our control policies will not have access to privileged simulation state information when deployed in the real world, we can still use privileged information to accelerate training in simulation. We use Asymmetric Actor Critic training~\cite{asymmetric_actor_critic}, in which our critic $V(\bm{s})$ is given all privileged state information $\bm{s}$ and our policy $\pi(\bm{o})$ is provided an observation $\bm{o}$, which is a limited subset of this privileged state information. With this method, the policy learns to perform the task using only observations we can capture in the real world, but the critic can leverage privileged state information to provide more accurate value estimates, improving the speed and quality of policy training. 

The state at timestep $t$ is $\bm{s}_t = [\bm{o}_t, \bm{v}_t, \bm{\omega}_t, t, \bm{f}^{\text{dof}}_t, \bm{F}^{\text{fingers}}_t]$,
where $\bm{o}_t $ is the observation, $\bm{v}_t \in \mathbb{R}^{3}$ is the object linear velocity, $\bm{\omega}_t \in \mathbb{R}^{3}$ is the object angular velocity, $t \in\mathbb{R}$ is current timestep, $\bm{f}^{\text{dof}}_t \in \mathbb{R}^{N^{\text{joints}}}$ is the vector of robot joint forces, $\bm{F}^{\text{fingers}}_t \in \mathbb{R}^{N^{\text{fingers}}}$ contains fingertip contact forces.

The training process utilizes a horizon length of 16 (i.e., the number of timesteps between updates for each robot, with all robots running in parallel) and 4096 parallel agents. The policy architecture consists of a multi-layer perceptron (MLP) with hidden layers of size [512, 512], an LSTM module with 1024 hidden units, and a critic network with hidden layers of size [1024, 512].

To improve the robustness and generalization of our policy, we apply extensive domain randomization during training. Randomizations are applied every 720 simulation steps and include variations in observations, actions, physics parameters, and object properties. Gaussian noise with a standard deviation of $0.01$ is added to both observations and actions. Gravity is perturbed additively using Gaussian noise with a standard deviation of $0.3$. The scale, mass, and friction of the object and table are randomized with a scaling parameter sampled from $[0.7, 1.3]$. The robot's scale, damping, stiffness, friction, and mass are also randomized with a scaling parameter sampled from $[0.7, 1.3]$.

We also introduce random force perturbations to the object. At each timestep, there is a 5\% probability of applying a force with a magnitude equal to 50 times the object's mass, directed along a randomly sampled unit vector. These perturbations serve two key purposes. First, they can displace the object before the robot makes contact, simulating real-world uncertainties such as unexpected disturbances or pose estimation errors. This encourages the policy to actively track and reach for objects that may not be precisely where they were initially observed. Second, if the object is already grasped, these perturbations can destabilize the grasp, promoting the development of robust and stable grasping strategies that minimize the likelihood of dropping the object.

All relevant code for PPO can be found \href{https://github.com/tylerlum/human2sim2robot/tree/main/human2sim2robot/ppo}{here}, and for simulation training found \href{https://github.com/tylerlum/human2sim2robot/tree/main/human2sim2robot/sim_training}{here}

\section{Sim-to-Real Inference-Time Details}
\label{app:real_time}

\subsection{Policy Inference-Time Details}

\begin{wrapfigure}{r}{0.6\textwidth}
\vspace{-0.4cm}
\centering
\includegraphics[width=\linewidth]{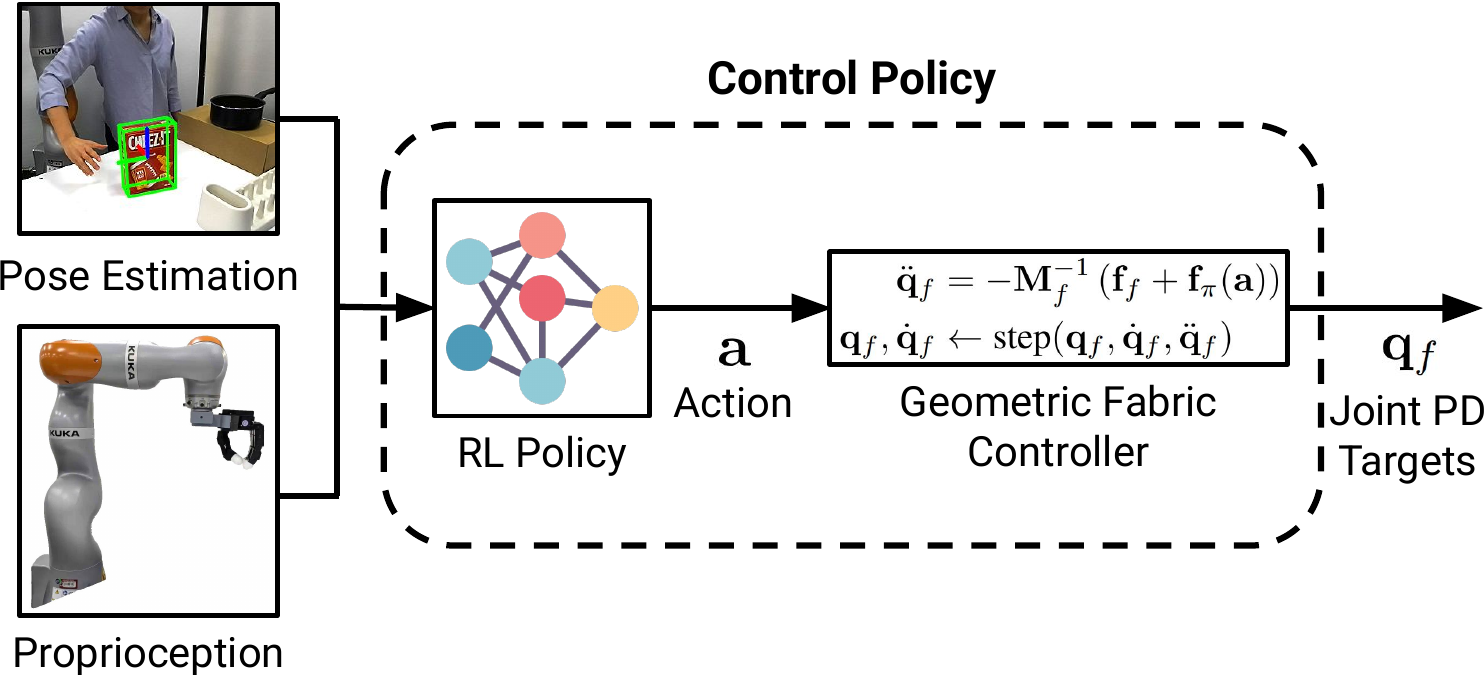}
\caption{\textbf{Inference-Time Diagram.} The policy takes object pose and robot proprio as input. It outputs an action that is sent to a geometric fabric controller~\cite{lum2024dextrahg}, which generates joint PD targets.}
\label{fig:inference_time_system_diagram}
\vspace{-0.4cm}
\end{wrapfigure}

Figure~\ref{fig:inference_time_system_diagram} illustrates the control pipeline at deployment, highlighting the inputs and outputs of our policy at test time. We track object 6D poses at 30Hz using FoundationPose~\cite{wen2024foundationpose}. The RL policy processes real-time observations and outputs actions at 15Hz, which are then passed to a geometric fabric controller~\cite{lum2024dextrahg} running at 60Hz. Finally, this controller produces robot joint PD targets, which are  executed by a low-level PD controller at 200Hz.

\subsection{Real-Time Perception Details}

In this section, we describe our real-time perception pipeline, which enables pose estimation at 30Hz using FoundationPose~\cite{wen2024foundationpose}. The process begins with pose registration to determine the object's initial pose, which takes approximately 1 second. This step requires a textured object mesh and a segmented object mask. To generate the mask, we pass the image and a text prompt into Grounding DINO~\cite{liu2023grounding} to obtain an object bounding box. The image and bounding box are then passed to SAM2~\cite{ravi2024sam2}, which produces a high-quality segmented object mask. The text prompt can either be manually provided or automatically generated by rendering the textured object mesh into an image and using GPT-4o to produce a descriptive prompt.

Once the initial pose is established, FoundationPose performs real-time object pose tracking by generating pose hypotheses near the previous estimate and selecting the best match, achieving a consistent 30Hz rate. Although tracking is generally robust, pose estimates can degrade when the object moves rapidly or becomes heavily occluded. To address this, we implement a separate pose evaluation process that monitors the pose estimate quality and triggers re-registration if needed.

This evaluation is performed by running SAM 2 at 1Hz to produce high-quality segmented object masks, which are treated as ground truth due to SAM 2's reliability, even under challenging conditions. The predicted object mask, generated using FoundationPose's pose estimate and the known camera parameters, is compared to the ground-truth mask using the intersection over union (IoU) metric. If the IoU falls below 0.1, FoundationPose reinitializes the pose registration process.

\section{Full Task List}
\label{app:tasks}

We perform experiments on the following tasks:

\begin{itemize}[leftmargin=1.5em]
    \item \textbf{\texttt{snackbox-push}}: The objective is to push the snackbox across the table until it makes contact with a static box. The snackbox is initialized face-down, with its position randomized within a 4 cm $\times$ 4 cm region. The task is considered successful if the snackbox contacts the static box.

    \item \textbf{\texttt{plate-push}}: Similar to \texttt{snackbox-push}, this task requires pushing a plate across the table until it contacts the static box. The plate starts in a flat orientation, with its position randomized within a 4 cm $\times$ 4 cm region. Success is achieved when the plate contacts the static box.

    \item \textbf{\texttt{snackbox-pivot}}: The goal is to pivot the snackbox from a face-down orientation to a sideways orientation using the static box for support. The snackbox is initialized in a face-down orientation, with its position randomized within a 1 cm $\times$ 4 cm region. Due to the robot’s initial position, the snackbox cannot be substantially moved in one direction. The task is successful if the snackbox is pivoted against the static box into a stable sideways orientation.

    \item \textbf{\texttt{pitcher-pour}}: This task involves grasping a pitcher by its handle, lifting it off the table, and reorienting it so that its spout is positioned above a static saucepan, simulating a pouring motion. The pitcher starts in an upright orientation, with its position randomized within a 4 cm $\times$ 4 cm region. The task is successful if the pitcher is lifted by its handle and correctly positioned with its spout above the saucepan.

    \item \textbf{\texttt{snackbox-push-pivot}}: This task combines the \texttt{snackbox-push} and \texttt{snackbox-pivot} tasks. The snackbox starts in a face-down orientation, with its position randomized within a 4 cm $\times$ 4 cm region. The task consists of two sequential steps. The push task is successful if the snackbox is first pushed to make contact with the static box, and the pivot task is successful if the snackbox is pivoted against the box into a sideways orientation. Success is graded on a three-level scale. The score is 0 if the push fails, 0.5 if the push is successful but the pivot fails, and 1 if both the push and pivot are successful.

    \item \textbf{\texttt{plate-lift-rack}}: The objective is to lift a plate and place it into a dishrack. The plate starts in an upright orientation, leaning against the static box, with its position randomized within a 0.5 cm $\times$ 4 cm region. Since the plate must remain leaning against the box, its movement is primarily constrained along the box length. The task consists of two sequential steps. The lift task is successful if the plate is lifted off of the table, and the rack task is successful if the plate is placed into the dishrack while maintaining an upright orientation. Success is graded on a three-level scale. The score is 0 if the lift fails, 0.5 if the lift is successful but the rack fails, and 1 if both the lift and rack are successful.

    \item \textbf{\texttt{plate-pivot-lift-rack}}: This task requires a sequence of actions: pivoting a plate against the static box, lifting it off the table, and placing it into a dishrack. The plate starts in a flat orientation next to the static box, with its position randomized within a 0.5 cm $\times$ 4 cm region.

    The task consists of three sequential steps. The pivot task is successful if the plate is pivoted against the static box to an upright orientation, the lift task is successful if the plate is lifted off of the table, and the rack task is successful if the plate is placed into the dishrack while maintaining an upright orientation. Success is graded on a four-level scale. The score is 0 if the pivot fails, 0.33 if the pivot is successful but the lift fails, 0.66 if the pivot and lift are successful but the rack fails, and 1 if the pivot, lift, and rack are successful.
\end{itemize}

\section{Baseline Details}
\label{app:baseline_details}

\subsection{Full Human Hand Trajectory Estimation}

Our baseline methods typically require robot action labels for every step of the task. When working with only a single human video demonstration, this can be done by estimating the human hand pose in each frame and retargeting it to the robot. Specifically, we perform hand pose estimation using HaMeR, followed by a depth alignment step (Appendix~\ref{app:hamer_alignment}). The resulting hand poses are mapped to robot joint configurations using an inverse kinematics (IK) procedure similar to that in Section~\ref{sec:human_demo_real_to_sim}.

However, simply solving the IK for each frame independently and stitching the results together often fails due to three main issues: errors in hand pose estimation, a lack of consistency/smoothness over time, and unreachable target poses. We address these issues as follows:

\begin{enumerate}[leftmargin=1.5em]
    \item \textbf{Mitigating Hand Pose Errors.} Hand pose estimation can suffer from large errors when fingers are occluded (Appendix~\ref{app:hamer_alignment}). To address this, we compare each newly estimated pose with the previous pose. If their distance exceeds a threshold, we skip the current frame’s pose rather than attempting to retarget an unreliable estimate.

    \item \textbf{Ensuring Joint Consistency.} To maintain smooth transitions between consecutive poses, we iteratively solve IK using the previous solution as both a default and a seed configuration. We employ cuRobo~\cite{curobo_report23} to generate 100 parallel solutions, each initialized by adding Gaussian noise (15$^\circ$ standard deviation) to the previous IK solution. We then select the solution whose joint angles have the smallest $\ell_{\infty}$ difference from the previous timestep’s configuration. If even this “best” solution differs excessively, we skip that frame.

    \item \textbf{Handling Unreachable Targets.} If the IK target is unreachable, we skip the corresponding frame.
\end{enumerate}

After applying these checks, we downsample the resulting trajectory and verify that all joint velocities remain below the robot’s limits. If any exceed the limit, we stretch the time between successive waypoints to reduce velocity. Although this procedure generally yields a plausible trajectory, inaccuracies can still arise in cases of severe finger occlusion or when the hand is far from the camera. In contrast, obtaining a reliable pre-manipulation hand pose is generally easier, as it requires only a single frame with accurate hand pose estimation. Such a frame is easier to find because the hand is typically not heavily occluded when it approaches the object, while the full demonstration can include much more occlusion of the hand.

\subsection{Replay Details}

In this section, we provide additional details on the implementation of replay. First, we need to clearly define the frames we care about. We define $ \bm{T}^{A\rightarrow B} $ as the relative transformation of frame $ B $ with respect to frame $ A $. This means $ \bm{T}^{A\rightarrow C} = \bm{T}^{A\rightarrow B}\bm{T}^{B\rightarrow C}$.

Let $R$ be the robot's base frame, $M$ be the robot's middle-finger frame, $C$ be the camera frame, and $O$ be the object frame. Given these four frames, we need three independent relative transforms to fully specify this system. The camera extrinsics calibration gives us $\bm{T}^{R\rightarrow C}$. FoundationPose object pose estimates give us $\bm{T}^{C\rightarrow O}$ at each timestep. HaMeR with depth refinement gives us hand pose estimates that allow us to compute the desired pose of the robot's middle-finger frame $\bm{T}^{C\rightarrow M}$. This fully specifies the demonstration. This allows us to compute the object trajectory $\{\bm{T}^{R\rightarrow O}_t\}_{t=1}^T$ and the robot's middle-finger trajectory $\{\bm{T}^{R\rightarrow M}_t\}_{t=1}^T$ for this demonstration. This is used to perform the inverse kinematics procedure described above to generate a robot configuration trajectory.

To execute replay in the real world, we can directly track this joint trajectory with joint PD control.

\subsection{Object-Aware Replay Details}
For object-aware replay, at timestep $t=1$, we use FoundationPose to estimate the new object pose $\bm{T}^{R\rightarrow O^{\text{new}}}_1$, which will be similar but not identical to the initial object pose during demonstration collection $\bm{T}^{R\rightarrow O}_1$. Our goal is to compute a new robot middle-finger trajectory $\{\bm{T}^{R\rightarrow M^{\text{new}}}_t\}_{t=1}^T$ that keeps the same relative pose between the object and the middle-finger as in the demonstration.

Let $\bm{T}^{O \rightarrow M}$ be the relative pose between the object and the middle-finger at each timestep of the demonstration. This can be computed as $\bm{T}^{O \rightarrow M} = (\bm{T}^{R \rightarrow O})^{-1} \bm{T}^{R \rightarrow M}$. Our goal is to maintain this same relative relationship during replay, such that $\bm{T}^{O \rightarrow M} = \bm{T}^{O^{\text{new}} \rightarrow M^{\text{new}}}$.

The new middle-finger trajectory can then be computed as:
\begin{align}
\bm{T}^{R \rightarrow M^{\text{new}}}_t &= \bm{T}^{R \rightarrow O^{\text{new}}}_t \bm{T}^{O^{\text{new}} \rightarrow M^{\text{new}}}_t \\
&= \bm{T}^{R \rightarrow O^{\text{new}}}_t \bm{T}^{O \rightarrow M}_t \\
&= \bm{T}^{R \rightarrow O^{\text{new}}}_t (\bm{T}^{R \rightarrow O}_t)^{-1} \bm{T}^{R \rightarrow M}_t
\end{align}

To compute $\bm{T}^{R \rightarrow O^{\text{new}}}_t$ for $t > 0$, we assume the object follows the same relative motion as in the demonstration, but starting from the new initial pose. This can be computed as:
\begin{align}
\bm{T}^{R \rightarrow O^{\text{new}}}_t &= \bm{T}^{R \rightarrow O^{\text{new}}}_1 (\bm{T}^{R \rightarrow O}_1)^{-1} \bm{T}^{R \rightarrow O}_t
\end{align}

Substituting this into our equation for the new middle-finger trajectory:
\begin{align}
\bm{T}^{R \rightarrow M^{\text{new}}}_t &= \bm{T}^{R \rightarrow O^{\text{new}}}_1 (\bm{T}^{R \rightarrow O}_1)^{-1} \bm{T}^{R \rightarrow O}_t (\bm{T}^{R \rightarrow O}_t)^{-1} \bm{T}^{R \rightarrow M}_t \\
&= \bm{T}^{R \rightarrow O^{\text{new}}}_1 (\bm{T}^{R \rightarrow O}_1)^{-1} \bm{T}^{R \rightarrow M}_t
\end{align}

This simplifies to:
\begin{align}
\bm{T}^{R \rightarrow M^{\text{new}}}_t &= \bm{T}_{\text{RELATIVE}} \bm{T}^{R \rightarrow M}_t
\end{align}

where $\bm{T}_{\text{RELATIVE}} = \bm{T}^{R \rightarrow O^{\text{new}}}_1 (\bm{T}^{R \rightarrow O}_1)^{-1} = \bm{T}^{R \rightarrow O}_1 \bm{T}^{O \rightarrow O^{\text{new}}}_1 (\bm{T}^{R \rightarrow O}_1)^{-1}$ is the transformation that accounts for the change in the initial object pose. This transformation is applied to the entire middle-finger trajectory, effectively adjusting the demonstration to the new initial object pose while preserving the relative motion between the object and the middle-finger.

To execute object-aware replay in the real world, we can directly track this new joint trajectory with joint PD control.

\subsection{Behavior Cloning Details}

To train a Diffusion Policy, we need to collect a dataset of observation and action pairs. Because we only have one human demonstration, we can generate additional demonstration data by introducing small amounts of transformation noise to the object's pose and then use the process above to compute robot joint configuration trajectories with adjusted IK targets that account for this noise. Specifically, we sample $\bm{T}^{O \rightarrow O^{\text{new}}}_1$ with the same translation and rotation noise as used in RL training and then run the object-aware replay computation above to get new object pose trajectories and robot joint configuration trajectories. The observation consists of the vector of robot joints $\bm{q}_t$, the palm pose $\bm{p}_t^{\text{palm}}$, and the object pose $\bm{p}_t^{\text{object}}$. The action consists of joint position targets relative to the current robot position. We train with batch size 128, 50 diffusion iterations, learning rate 1e-4, and weight decay 1e-6. We use the state-based diffusion policy implementation from~\citet{drolet2024}.

\section{Qualitative Analysis}
\label{app:qualitative_analysis}

\begin{figure}
    \centering
    \includegraphics[width=\linewidth]{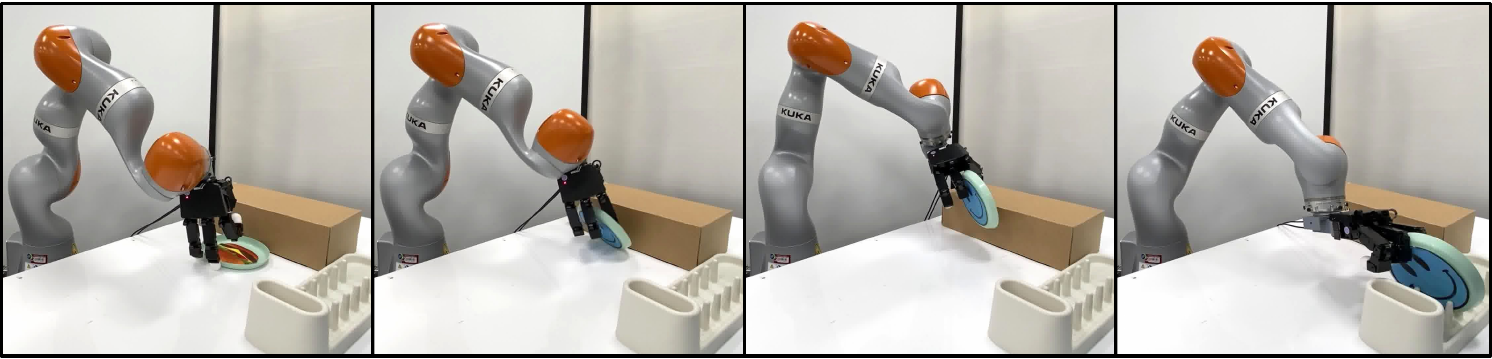}
    
    \caption{\textbf{Plate Pivot Lift Rack.} Robot converges on a strategy that is guided by the human demonstration, but adapted to its morphological differences.}
    \label{fig:optimal_policy}
\end{figure}

Qualitatively, for tasks like \texttt{plate-pivot-lift-rack} that are more intricate, small differences in the pre-manipulation hand pose can result in very different learned strategies due to the differences in the human and robot morphologies. For instance, while the human hand used the pinky and ring fingers to lift the plate before transitioning to a grasp, the Allegro hand, which is much larger, used its ring finger to pivot the plate and clipped it between the middle and index finger once the plate was off the table (see Figure~\ref{fig:optimal_policy}). This further underscores our hypothesis that significant differences in robot morphologies may lead to strategies that are guided by the human motion but ultimately converge on a different strategy that is more suitable for the robot's embodiment after learning through trial-and-error. 

Regarding \textsc{Human2Sim2Robot} failure modes, failures typically arose from converging on policies that exploited simulation inaccuracies or from significant pose estimation error from occlusion. Examples of simulation inaccuracies include imperfect friction modeling of the tabletop and static objects, such that policies converged on behavior that leveraged these inaccurate parameters. The most challenging task was \texttt{plate-pivot-lift-rack} as it required high precision object handling: the plate is very thin, and is hard to manipulate and slips out of the large Allegro hand easily.

The policy currently needs to be initialized in the rough region of the pre-manipulation pose. To eliminate the need to initialize the robot hand in close proximity to the object, methods such as collision-free motion planning or an initial approach stage reward formulation (like having a reward for minimizing the L2 distance to the pre-manipulation pose) have potential to overcome this challenge. These techniques can be seamlessly integrated into our system to facilitate navigation from a default rest pose to the pre-manipulation pose. Our experiments and ablations here highlight the significant effectiveness of the pre-manipulation pose in guiding the RL policy toward learning human-like behaviors for contact-rich manipulation tasks.

\section{Other Embodiments}
\label{app:other_embodiments}

\begin{figure}
    \centering
    \includegraphics[width=\linewidth]{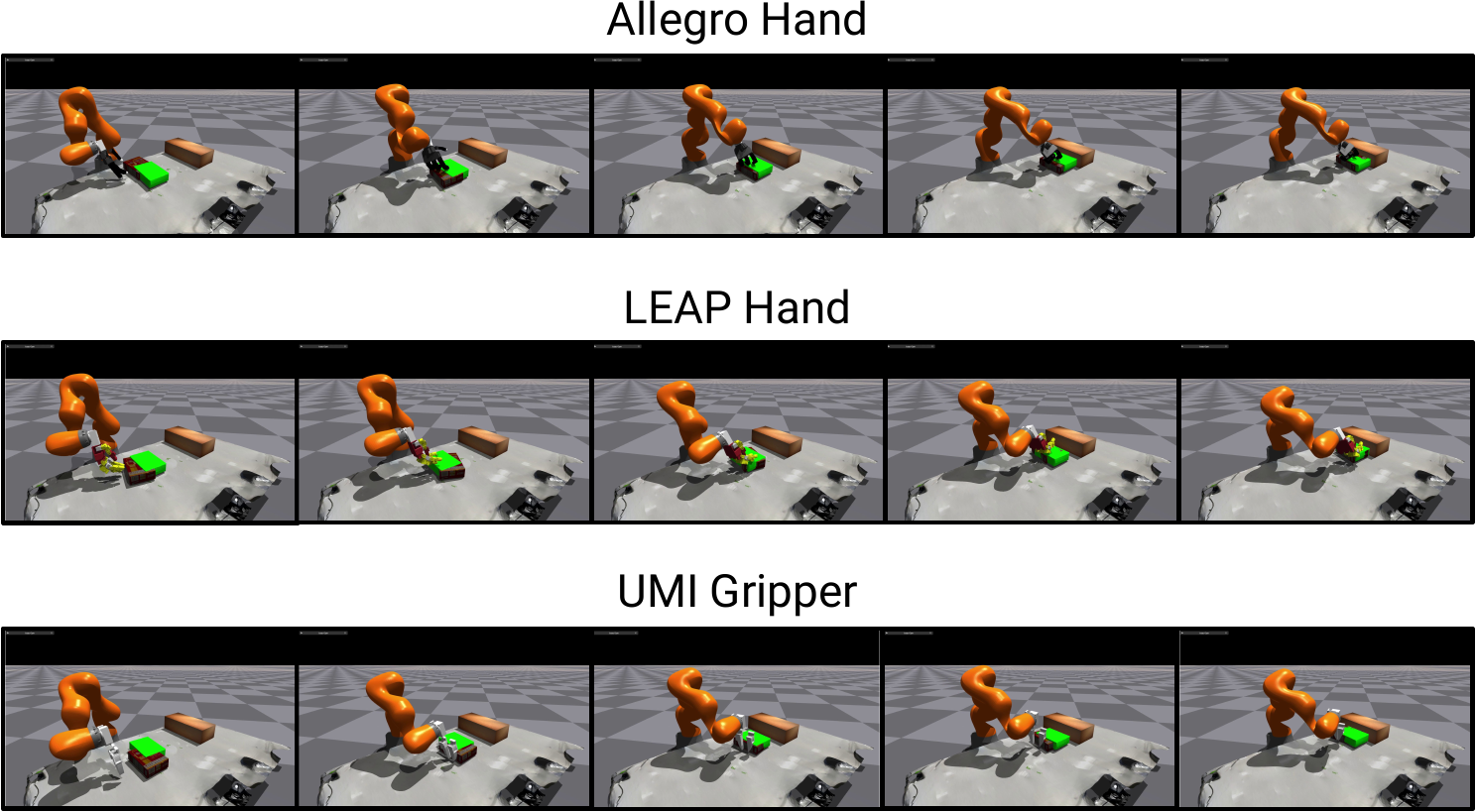}
    \caption{\textbf{Other Embodiments.} \textsc{Human2Sim2Robot} can be applied to different robot embodiments, including an Allegro Hand, a LEAP Hand~\cite{shaw2023leaphand} and a UMI gripper~\cite{chi2024universal}. This is demonstrated with preliminary simulation experiments for the \texttt{snackbox-push} task. The green box represents the target pose of the snackbox.}
    \label{fig:other_embodiments}
\end{figure}

Although our experiments are primarily tested on a Kuka arm and Allegro hand, we expect \textsc{Human2Sim2Robot} to work on other robot embodiments, as there are no aspects of the framework that are specific to this embodiment. To validate this, we perform initial experiments demonstrating \textsc{Human2Sim2Robot} on a LEAP Hand~\cite{shaw2023leaphand} and UMI gripper~\cite{chi2024universal} on the \texttt{snackbox-push} task. Figure~\ref{fig:other_embodiments} shows qualitative results using these embodiments, which shows that this was successful without any reward tuning. The only modifications required to make this work were changing the robot URDF and the robot configuration for retargeting via inverse kinematics (changing link names, relative orientations, default joint configuration, collision spheres), geometric fabric controller (link names, default joint configuration, collision spheres), and simulation environment (link names, action/observation dimensions). 

\section{Additional Related Work}
\label{app:additional_related_work}

\textbf{Visuomotor Imitation Learning for Robotics.} Recent work in visuomotor IL for robotic manipulation has shown success learning from a large number of expert demonstrations~\cite{rt22023arxiv, rt12022, jang2021bc, kim24openvla,lin2024datascalinglawsimitation,wang2024dexcap,chi2024diffusion,zhao2024aloha}. Demonstrations are typically collected through teleoperation or specialized wearable equipment~\cite{chi2024universal,lin2024datascalinglawsimitation,wang2024dexcap} like motion capture gloves, AR/VR equipment, or portable robot hands, which makes scaling of data collection efforts expensive. In contrast, videos of human demonstrations are inexpensive to collect and more intuitive to demonstrators. Since videos lack explicit action labels, a popular approach is to obtain per-timestep human hand pose estimates and convert them into robot action labels through IK-based retargeting~\cite{wang2024dexcap,chen2024arcap}. However, the human-robot embodiment gap often makes finding an IK solution infeasible or result in a robot joint configuration that is suboptimal for reproducing the demonstrated task. This poses significant challenges for visuomotor IL methods that directly rely on accurate correspondences between the demonstrated and learned behaviors.

Therefore, instead of directly learning a mapping from observations to actions, \textsc{Human2Sim2Robot} performs RL in simulation guided by a single human video demonstration. This approach acknowledges that while human demonstrations provide useful guiding strategies for completing the task, certain actions may not be suitable for robots given substantial embodiment differences. Learning dexterous manipulation policies with RL encourages human-like behavior when beneficial while allowing deviations when the human strategy is unsuitable for the robot's embodiment. Furthermore, robust recovery and retry behavior emerges as a result of training and does not need to be explicitly demonstrated as in IL.

\textbf{One-Shot Imitation Learning.}
Another category of IL methods that parallels our approach follows the one-shot IL (OSIL) paradigm, where a single demonstration is provided. Past work has performed object pose estimation~\cite{heppert2024ditto, vitiello2023one} or object-aware retargeting using open-world vision models~\cite{okami2024} to transfer the demonstrated trajectory to novel scenes, or adapted the single demonstration to a new scene by leveraging object segmentation and visual servoing to move the robot's end effector to the same position relative to the object before replaying the demonstration~\cite{valassakis2022dome}. Beyond strictly using one demonstration, a related approach collects a small number of ($\sim$5) teleoperated demonstrations per task, then generates more data by retargeting demonstrations to new initial conditions and rolling out these trajectories in a simulated digital twin, executing these actions in the real world if they successfully complete the task in simulation~\cite{jiang2024dexmimicgen}. 

While these approaches are more data-efficient than visuomotor IL policies, they suffer from limited generalization beyond the demonstrated actions. Furthermore, when learning from human video demonstrations, object and hand pose estimates are often inaccurate due to occlusions and noise. Even if pose estimates are accurate, the human-robot embodiment gap causes IK-based retargeting to produce infeasible robot hand trajectories. Therefore, simply replaying modified versions of the single demonstration or directly learning observation-to-action mappings from retargeted data is unlikely to succeed. Our insight is that using the single human video demonstration to provide task specification and guidance for RL more effectively leverages this data source for robot learning. This allows robots to develop effective strategies with their own embodiment, rather than rigidly imitating human behaviors.

\textbf{Reinforcement Learning for Robotics.} Reinforcement learning (RL) is a method for training autonomous agents to perform complex tasks by interacting with an environment through trial and error. In this work, we train manipulation policies in simulation to avoid the pitfalls of RL in the real world such as slow training, unsafe behavior, frequent environment resets, and difficult to tune reward functions~\cite{torne2024rialto, zhu2020ingredientsrealworldroboticreinforcement}. Sim-to-real RL has shown significant promise across a wide range of robotic tasks, achieving state-of-the-art performance in domains such as legged locomotion~\cite{cheng2023parkour, margolisyang2022rapid, miki2022quadruped}, drone racing~\cite{kaufmann2023champion}, bipedal soccer~\cite{haarnoja2024soccer}, and in-hand manipulation~\cite{qi2022hand}. However, this potential has yet to be fully realized for dexterous manipulation over the full robot arm-and-hand kinematics: many prior works on RL for dexterous manipulation are confined to simulation with non-physical, floating-hand robots~\cite{chen2022towards, Rajeswaran-RSS-18, wan2023unidexgrasp++, xu2023unidexgrasp}. 

Recent works leveraging sim-to-real RL for real-world manipulation tasks have used RL to fine-tune a BC policy~\cite{torne2024rialto} or learn a residual policy that outputs delta actions relative to an open-loop base motion~\cite{chenobject}. Fine-tuning a BC policy may learn more stably, but requires a demonstration dataset with the same robot embodiment. Residual policy learning can be effective, but only allows small adjustments to the base motion, limiting its ability to overcome large embodiment gaps or demonstrate retry behavior. Additionally, both methods require accurate human action sequences from high-quality motion capture or teleoperation. \textsc{Human2Sim2Robot} is a real-to-sim-to-real framework that trains an RL policy over a full arm-and-hand action space, guided by object-centric rewards and a pre-manipulation hand pose from just one human demonstration. This approach can thus accommodate lower-quality human video demonstration data, facilitate learning of retry behavior, and overcome the human-robot embodiment gap. Furthermore, our approach enables policies to learn contact-rich dexterous manipulation tasks that are more complex than parallel-jaw pick-and-place or in-hand manipulation with a static arm. 

\citet{lum2024dextrahg} trains dexterous grasping policies using sim-to-real RL with a geometric fabric controller, enabling smooth, coordinated motion while effectively avoiding undesired collisions with the environment. However, the policy is limited to a simple grasping task, typically converging on a simple top-down grasp as the hand is initialized above the object and the policy is trained from scratch without human guidance. \textsc{Human2Sim2Robot} leverages a similar geometric fabric controller, but enables policies to perform more general prehensile and non-prehensile manipulation tasks by incorporating human guidance on how to perform the task.

Recent studies corroborate that pre-grasp hand configurations can accelerate policy learning and result in human-like grasps~\cite{Ciocarlie2007DexterousGV, dasari2023pgdm, luo2024graspingdiverseobjectssimulated}. These methods only focus on grasping tasks, use pre-grasps from very similar embodiments, and have only been tested in simulation. In this work, we focus on training RL policies that transfer to the real-world, overcome the human-robot embodiment gap, and perform both prehensile and non-prehensile manipulation. Furthermore, we incorporate scalable methods for pre-manipulation hand pose acquisition from one human video demonstration.

\end{document}